\documentclass[runningheads]{llncs}
\usepackage[T1]{fontenc}
\usepackage{bbding}
\usepackage{framed,multirow}
\usepackage{amssymb}
\usepackage{latexsym}
\usepackage{xcolor}
\usepackage{makecell}
\usepackage{lineno,hyperref}
\modulolinenumbers[5]
\usepackage{amsmath}
\usepackage{verbatim}
\usepackage{diagbox}
\usepackage{multirow}
\usepackage{color,graphicx}
\usepackage{arydshln}
\usepackage{epstopdf}
\usepackage{hyperref}
\usepackage{url}
\usepackage{footnote}
\usepackage{bm}
\usepackage{cite}
\usepackage{subfigure}
\usepackage{verbatim}
\usepackage{ulem}
\usepackage[ruled,linesnumbered]{algorithm2e}

\definecolor{Mycolor1}{HTML}{036564}
\definecolor{Mycolor2}{HTML}{DE7D2C}

\usepackage{setspace} 

\usepackage{enumitem}

\begin{document}

\title{Multi-Target Landmark Detection with Incomplete  Images via Reinforcement Learning and Shape Prior Embedding\thanks{www.sdspeople.fudan.edu.cn/zhuangxiahai/}}

\author{Kaiwen Wan\inst{1} \and
Lei Li\inst{2} \and 
Dengqiang Jiar\inst{3} \and
Shangqi Gao\inst{1} \and
Wei Qian\inst{4} \and
Yingzhi Wu\inst{5} \and
Huandong Lin \inst{6} \and
Xiongzheng Mu \inst{5} \and
Xin Gao \inst{6} \and
Sijia Wang \inst{4} \and
Fuping Wu \inst{7} \and
Xiahai Zhuang* \inst{1}}
\authorrunning{Kaiwen Wan et~al.}

\institute{School of Data Science, Fudan University, Shanghai, 200433, China \and
Institute of Biomedical Engineering, University of Oxford, Oxford, UK \and
School of Naval Architecture, Ocean and Civil Engineering, Shanghai Jiao Tong University, Shanghai, China \and
Shanghai Institute of Nutrition and Health, University of Chinese Academy of Sciences, Chinese Academy of Sciences, Shanghai 200031, China \and
Department of Plastic Surgery, Huashan Hospital, Fudan University, Shanghai 200040, China. \and
Department of Endocrinology and Metabolism, Zhong Shan Hospital, Fudan University, 200032 Shanghai, China \and
Nuffield Department of Population Health, University of Oxford, Oxford, UK}

\maketitle
\begin{abstract}
Medical images are generally acquired with limited field-of-view (FOV),  
which could lead to \textit{incomplete} regions of interest (ROI), and thus impose a great challenge on medical image analysis.
This is particularly evident for the learning-based multi-target landmark detection,
where algorithms could be misleading to learn primarily the variation of background due to the varying FOV, failing the detection of targets. 
Based on learning a navigation policy, instead of predicting targets directly, reinforcement learning (RL)-based methods have the potential to tackle this challenge in an efficient manner.
Inspired by this, in this work we propose a multi-agent RL framework for simultaneous multi-target landmark detection. 
This framework is aimed to learn from incomplete or (and) complete images to form an implicit knowledge of global structure, which is consolidated during the training stage for the detection of targets from either complete or incomplete test images.  
To further explicitly exploit the global structural information from incomplete images, we propose to embed a shape model into the RL process.  
With this prior knowledge, the proposed RL model can not only localize dozens of targets simultaneously, but also work effectively and robustly in the presence of incomplete images.
We validated the applicability and efficacy of the proposed method on various multi-target detection tasks with incomplete images from practical clinics, using body dual-energy X-ray absorptiometry (DXA), cardiac MRI and head CT datasets.  
Results showed that our method could predict whole set of landmarks with incomplete training images up to 80\% missing proportion (average distance error 2.29 cm on body DXA),
and could detect unseen landmarks in regions with missing image information outside FOV of target images (average distance error  6.84 mm on 3D half-head CT).   
Our code will be released via
\href{https://zmiclab.github.io/projects.html}{https://zmiclab.github.io/projects.html}, once the manuscript is accepted for publication.


\end{abstract}

\begin{spacing}{1.00} 
\section{Introduction}
Landmark detection is an essential step for biomedical image analysis \cite{ZHAN201645}. Anatomical landmarks can describe morphological characteristics of anatomical structures, which is important in morphometric and medical analysis. 
Their locations can also be useful in many downstream computing tasks, such as image segmentation and registration \cite{Seghers2007LandmarkBL}. 
However, in practical clinics, medical images usually have varying field-of-view (FOV) and cover incomplete region-of-interests (ROI).
Such incomplete images lead to particular challenges for learning-based landmark detection.

\begin{figure*}[!t]
\centering
\includegraphics[width=0.95\textwidth]{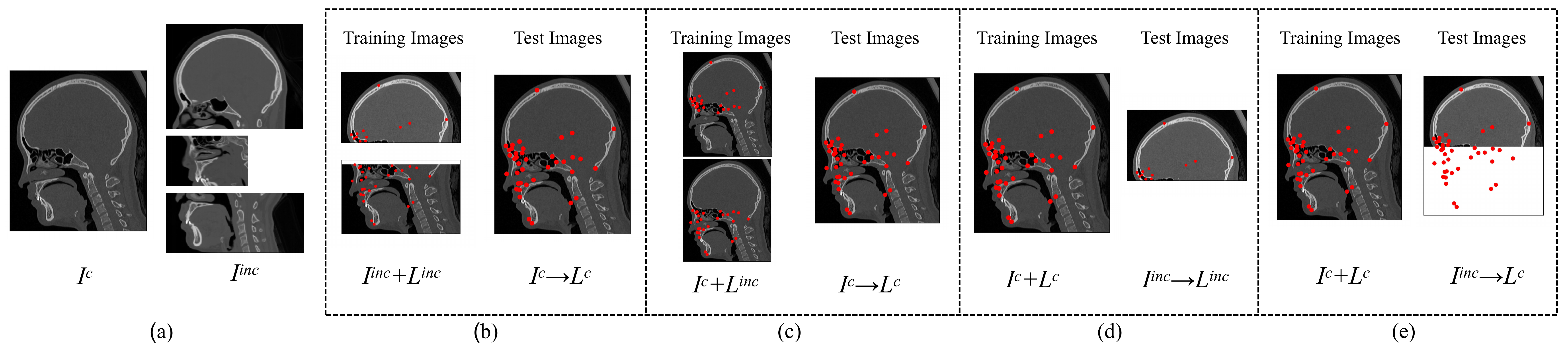}
\caption{\label{fig:fig_intro}
Illustration of complex scenarios involving incomplete images for machine learning. Here, $I^c$ ($L^c$) and $I^{inc}$ ($L^{inc}$) respectively refer to complete images (labels) and incomplete images (labels);  the dots (\textcolor{red}{$\bullet$}) represent landmarks. 
(a) a complete image and three incomplete images of head CT in saggital view; 
(b) a scenario where the machine learning algorithm is required to learn from incomplete images (with incomplete gold standard labels) to infer all completes label inside of the complete test image
;
(c) a scenario where the training images are complete but their gold standard labels are incomplete, and the target is to infer all complete label;
(d) and (e) scenarios where training images are complete in both images and labels, but the test images are incomplete. 
Differently, in (d) the target is to infer only incomplete label, \textit{i.e.,} landmarks, from the incomplete image given the available ROI, while in (e) the task is to predict all landmarks even for the ones not contained in the incomplete image.  
Note that images in this figure are all 2D saggital view of 3D images, and the landmarks on them are mapped from 3D space.
}
\end{figure*}

We refer an image as \textit{incomplete image} when it can not cover the whole ROI containing all targets, or when it, as a training image, does not provide gold standard labels of all targets.  
Fig.~\ref{fig:fig_intro} (a) illustrates the saggital view  of a complete CT image which covers all the anatomical landmarks of the head, and three incomplete images which only cover parts of the head. 
This incompleteness of medical images is common in clinical practice. 
For example,
many of the CT images are acquired without a whole-head scan in clinic to reduce exposure to radiation, such as mandible modeling 
\cite{app12031358},
nasal cancer study 
\cite{li2019tumor} and
stroke diagnosis
\cite{pmid33957774}.

These images become incomplete when the studies involve the target landmarks of the whole head, namely a complete image in one study can become incomplete in another for wider ROI.
This is even more common in dual-energy X-ray absorptiometry (DXA) scans, which are widely used to analyze \textit{different} areas of the body. 
These DXA images from different studies are incomplete for the study whose targets involve the whole body \cite{BRIOT2013265}. 
Finally, partial labeling is commonly seen in big medical image training data, since manually labelling is tedious and difficult even for experienced annotators.
Moreover, due to the difference of targets in different studies, the collected label can be partial even for a complete image \cite{fries2019weakly}.
  
Fig. \ref{fig:fig_intro} summarizes and illustrates the common scenarios involving incomplete images in multi-target landmark detection. 
Particularly, Fig. \ref{fig:fig_intro}~(b) shows the most common scenario, where the training images contain only limited and incomplete ROI, consequently their labels are incomplete; but the target is to detect all landmarks given a complete image.  
This is common yet challenging, and the core is to learn the global structural information (also known as shape information) from limited FOV images, with an effective scheme to consolidate the global knowledge for inference.  
Fig. \ref{fig:fig_intro}~(c) represents a similar yet different application, where images are all complete, in both of the training and test stages; but the gold standard labels for training are incomplete. 
This can happen in multi-center studies, where different center has interests of different targets. The ultimate goal of machine learning here is to learn  all the  knowledge of local annotations, and assemble them into a global model for inference. 
Fig.~\ref{fig:fig_intro} (d) and (e) demonstrate the more practical situations, where our training images are perfectly complete. 
However, in clinic practice the acquired images can be of limited FOV and incomplete. 
It is therefore desirable that the artificial intelligent models can be robust in the presence of such incomplete test images. 
Furthermore, in certain extreme cases the missing part of the incomplete test images can be so important that we need to recover these missing landmarks, as Fig.~\ref{fig:fig_intro} (e) shows, the  model is then expected to predict the targets even though they are not within the FOV of the test images.

The aforementioned scenarios can challenge the learning-based multi-target landmark detection algorithms. 
With incomplete images, the global structural information is not consistently presented, thus the algorithms may not be able to assemble the partial knowledge into global one.  
In addition, one algorithm developed for coping with incomplete training images may not be applicable to the scenario of incomplete test images, or vice versa. 
For deep learning-based methods, many models even require fixed sizes of input and output images, thus they can not be directly extended for such incomplete images with random FOVs and sizes.


Reinforcement learning (RL),
using an intelligent agent to study interacting with changing environments,
has the potential of tackling the challenges \cite{DBLP}.
The agent in RL can be trained to learn from the feedback for its own actions and experiences.
As a special type of machine learning methods, RL has been applied to numerous tasks in medical image analysis \cite{ZHOU2021102193}, including  object/ lesion detection
\cite{10.1007/978-3-319-66179-7_76}, registration \cite{Liao_Miao_de}, view plane localization \cite{Alansary2018AutomaticVP} and landmark detection \cite{ALANSARY2019156}.
Different from the conventional manner of learning and predicting targets directly, RL tries to learn a \textit{navigation policy} with an artificial agent in landmark detection. 
Such agent can be trained to search for optimal trajectories to target landmark from any random initial position of the image. 
This patch-based strategy captures local image information more easily than the holistic image information.
In the presence of incomplete images, RL-based algorithms based on such local information can robustly avoid disturbance caused by missing structures.
Hence, RL has the potential of offering a solution for the learning and inference with incomplete images, thanks to the navigation strategy for identifying the global structure of incomplete information.



For incomplete images, this work further
aims to develop a robust opportunity RL algorithm to detect multiple targets simultaneously.
Potentially, RL with multiple agents can search for multiple targets and accomplish the mission simultaneously, with each agent targeting a single landmark.
However, since each agent is navigating separately, this strategy may not fully utilize the global structural information, leading to additional challenges in scenarios of incomplete images.
Embedding prior shape information into RL-based algorithms can be a solution. 
For example, \cite{10.1007/978-3-030-59725-2_38} tried to add reward based on shape information to share them between different agents.
\cite{ghesu2018towards} used extra statistical shape modeling and robust estimation theory containing shape information to correct the initial prediction of RL.
Nevertheless, existing studies mainly focus on dealing one scenario at one time, and a unified framework applicable to these general problems of incomplete images is yet to be researched and developed.
Different scenarios may present different challenges for different methodologies, as  Fig.~\ref{fig:fig_intro} shows.  

In this work, we propose a two-stage deep neural network model for multi-target landmark detection with incomplete images. 
This model combines a multi-agent RL network and a statistical shape prior embedded network.  
The former network, extended from the single-agent RL framework, is aimed to search for multiple targets simultaneously from incomplete images; 
the latter, to provide prior knowledge of global structure for the former, is developed based on a statistical shape model.
The two networks are trained alternately, 
and the output of RL network can provide more samples for the shape network during training, which enhances the robustness of the landmark detection model.
Similar alternation is also applied in the inference (test) stage. 


The rest of this work is organized as follows: 
Sec.~\ref{related} presents the related work.
We elaborate on the methodologies in details in Sec.~\ref{method}.
Sec.~\ref{exp} provides the validation work, with comprehensive experiments and results via five practical applications from clinic perspectives and four datasets. 
Finally, we conclude and discuss in Sec.~\ref{conclusion}.

\section{Related Work}\label{related}
\subsection{Landmark detection algorithms}
Most conventional landmark detection methods process only complete images in different ways to exploit the structural knowledge.
Several works tried to build a global structure containing the explicit relationship between landmarks. 
For example, \cite{cootes1995active} proposed a statistical model to fit the object in target images, namely active shape model (ASM). 
ASM was later improved by adding global or local appearance information  \cite{927467,Cristinacce2006FeatureDA}.
Methods could also embed the shape information implicitly via regression or registration.
For example, regression maps image appearance to landmark locations directly \cite{6247976,6553766}. 
In contrast, registration builds a map between different images and it can propagate label information from atlas images to the target \cite{10.1007/978-3-319-13972-2_16,journal/MedIA/li2020}. These traditional methods can obtain promising results, but are usually time expensive.

For landmark detection, there is a trend to shift from traditional methods to deep learning-based methods.
Deep learning-based methods are much more efficient and can achieve results that may not be worse than traditional methods \cite{journal/MedIA/zhuang2019}. 
Using powerful deep neural networks to extract features, direct regression methods can predict the landmark locations in the target image with one step
\cite{6619290,6755923,zhang2016joint}.
Cascaded regression models were also proposed, which found coarse predictions at the beginning, and gradually updated localization in the following stages \cite{zheng20153d,lv2017deep,he2021cephalometric}.
However, most deep learning-based methods struggle to handle high-dimensional image data \cite{pmid34471202} and may be sensitive to the bounding box of images that cover the ROI \cite{SAGONAS20163}.

RL can effectively avoid the above problems by inputting sequential patches based on Markov decision process. 
Instead of locating the target directly, RL tried to learn the optimal path to the target from a random position with an agent.
\cite{10.1007/978-3-319-46726-9_27} adopted a deep RL agent to navigate in a 3D medical image for automatic landmark detection.
This method was improved with a multi-scale strategy later \cite{ghesu2017multi}.
To detect multiple targets simultaneously, \cite{vlontzos2019multiple} extended the RL network to be able to train multiple agents at the same time.
The structural relationship between landmarks has been considered in several methods for robust detection. 
For example, \cite{leroy2020communicative} proposed an RL scheme that used communicative agents to share information.
\cite{10.1007/978-3-030-59725-2_38} proposed an algorithm incorporating priors on physical structure, where graph communication layers and additional rewards were designed to further exploit structural priors.
Besides, \cite{10.1007/978-3-319-46726-9_27} also concluded that RL can judge whether the anatomical landmark is absent, which illustrated RL has the potential for incomplete image scenarios.


\subsection{Incomplete images}
The use of incomplete image data has been an ongoing research direction in recent years \cite{4711880}.
Traditional methods of processing incomplete images were usually based on image completion \cite{899371,899372,1038991}, in which constructed images without prior knowledge of their contents and could lead to unreliable results.
There are also traditional algorithms that can be applied directly to incomplete image data without using image reconstruction, such as using space variant operators to extract landmarks 
\cite{4711880}.
However, the incomplete images they studied were regularly or irregularly sampled from complete images, where missing pixels were distributed over the whole image.
These images were different from the incomplete medical images discussed in this work, which were caused by limited ROI.

Efforts have been devoted to deep learning-based methods for landmark detection of incomplete images. 
For instance, \cite{wang2020robust} used a random mask-based data augmentation strategy to generate incomplete images for network training.
\cite{HANAOKA2017192} proposed a two-stage sampling algorithm that can be applied to incomplete CT images.
As mentioned before, RL can detect absent anatomical landmarks and has also been used to handle incomplete images.
\cite{jain2020robust} proposed an RL-based multi-target landmark detection method, which could estimate primary target landmark locations even though the local images around targets are defaced.
\cite{ghesu2018towards} used a two-stage model, first training artificial agents to search for anatomical structures and then using elements of a statistical shape model to ensure the spatial coherence of the observed anatomical landmarks.
However, most of these studies focus on prediction with incomplete images given complete images for training, which is solely one scenario of the incomplete image problems, as illustrated in Fig.~\ref{fig:fig_intro} (d).
The complex nature and sophisticated scenarios involving incomplete images have not been fully explored yet. 
In contrast, this is the first study that aims to tackle this task, to the best of our knowledge.  

\section{Methodology}\label{method} 

\begin{table*} [t] \center
        \caption{Definition of symbols used in this paper.}
\label{tb:notion}
\resizebox{0.9\textwidth}{!}{
\begin{tabular}{c l|c l} 
\hline
Symbol & Definition &Symbol & Definition\\
\hline
$I_i$ & an image with index $i$&$\mathcal{J}_i$ & available index set of $I_i$ \\
$I_i^{c}/I_i^{inc}$ & a complete/incomplete image with index $i$   &$S$ & current state \\
$I^{c}/I^{inc}$ & scenarios involving complete/incomplete images    &$s_j$ & state of agent $j$\\
$L^{c}/L^{inc}$ & scenarios involving complete/incomplete label of images & $R$ &reward\\
$L^{c-inc}$ & scenarios involving complementary label of images & $r_j$&reward got by agent $j$\\
$N/N^{'}$ & number of training/test images &$A$ &current actions \\
$N_{dim}$ & dimension of the images&$a_j$ & action taken by agent $j$\\
$\mathbf X$ &landmark set library& ${M}_\mathrm{direction}$ & movement with given direction \\
$X_i$ &landmark set of $I_i$ &
$\Gamma_j^{[t]}$ & patch centered on position of agent $j$ at time $t$ \\
$\overline{X}$ & mean shape of  $\mathbf X$&$\epsilon$  &greedy parameter \\
$\hat{X}_i$ & prediction of $X_i$&$D$ & distance metric\\
$x_{ij}$ & landmark of $X_i$ with index $j$&
$x_j^d$ & ground-true location for agent $j$\\
$\hat{x}_{ij}$ & prediction of $x_{ij}$
&$x_j^{[t]}$& the locations of agent $j$ at time $t$\\
$\mathcal{J}$ &index set of all target landmarks&&\\
\hline
\end{tabular} }\\
\end{table*}

\begin{figure*}
\centering
\includegraphics[width=1\textwidth]{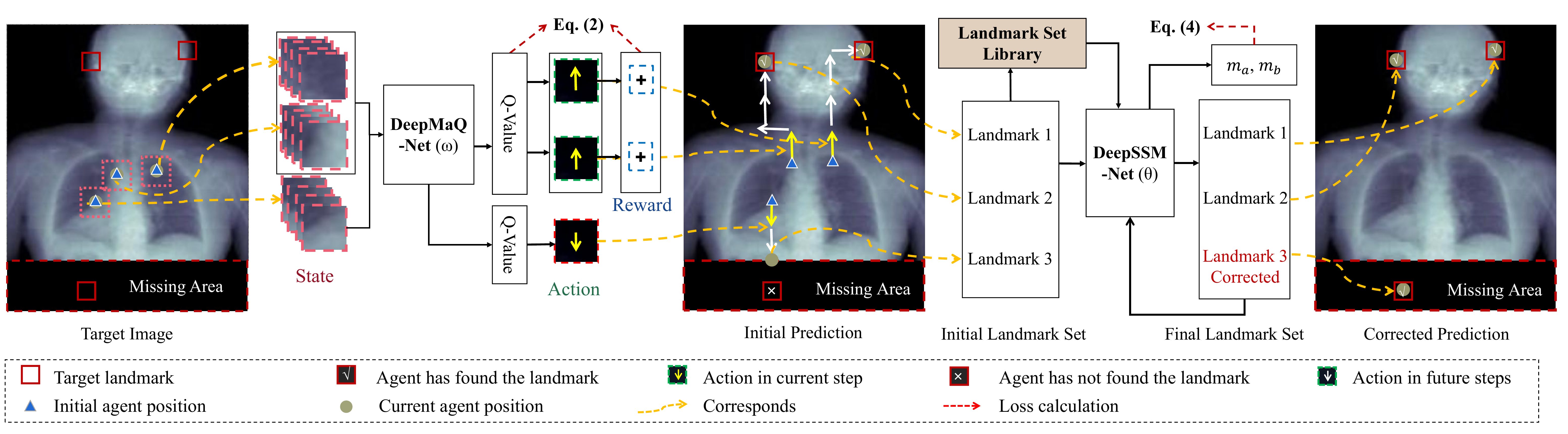}
\caption{\label{fig:fig2}Illustration of the shape-guided multi-agent reinforcement learning (SGMaRL) via DeepMaQ-Net and DeepSSM-Net. Here, we use three target landmarks for illustration, where one of them locates in the missing area. 
In the training stage, only agents whose targets exist in the image take actions and affect the calculation of loss function of DeepMaQ;
while in the test stage, all agents navigate with shape guidance and regularization via DeepSSM-Net.}
\end{figure*}

This work is aimed to develop an effective method for multi-target landmark detection in complex scenarios involving incomplete images.
In practice, the images collected either for training or test could be presented in a nonstandard form. 
Specifically, they may not cover the whole ROI, so certain volumes containing target landmarks could be missing.
These scenarios introduce the issue of unfixed number of landmarks, which challenges most of the deep learning-based approaches requiring predefined number of landmarks.

Let $\{(I_i^{c}, X_i,\mathcal{J})\vert _{i=1,\cdots, N}\}$ be a training set for a common task of landmark detection, where $\mathcal{J}$ is the index set of all target landmarks for a complete image such as $I_i^{c}$ whose corresponding landmark set is $X_i$. 
By contrast, in the scenario of incomplete images  a training set is denoted by $\{(I_{i}^{inc},X_i,\mathcal{J}_i)\vert _{i={1,\cdots, N}}\}$,  where $I_{i}^{inc}$ represents an incomplete image. The set of available landmark index for $I_{i}^{inc}$ is denoted as  $\mathcal{J}_i$, which is a subset of $\mathcal{J}$.

With similar notation, let $\{I_{i}^{inc}\vert _{i=1,\cdots, N^{'}}\}$ be the test image set with incomplete images. A model, denoted as $F$, predicts the coordinates of all landmarks, \textit{i.e.,} $\hat{X}_{i} = F(I_{i}^{inc}) \in \mathbb{R}^{\vert \mathcal{J}\vert \times {N_{dim}}}$, where $N_{dim}$ is dimension of the images. 
Intuitively, a workable model should have the following minimization problem:
\begin{equation} 
 \min\sum_{i=1}^{N^{'}} \sum_{j\in{\mathcal{J}_{i}}}D(x_{ij},\hat{x}_{ij}),  
\end{equation} 
where $x_{ij}\in \mathbb{R}^{N_{dim}}$ and $\hat{x}_{ij}\in \mathbb{R}^{N_{dim}}$ are respectively the ground truth and prediction of coordinate for the landmark with index $j$ in image $I_i^{inc}$;
$D(\cdot,\cdot)$ is a distance metric, such as the Euclidean distance.


To fulfill the above objective, we propose a two-stage deep neural network based on shape-guided multi-agent 
reinforcement learning (SGMaRL). 
As illustrated in Fig. \ref{fig:fig2}, 
this framework consists of two major components, \textit{i.e.,} a deep multi-agent Q-learning (DeepMaQ) network for multi-target landmark detection, and a deep statistical shape model (DeepSSM) for shape guidance and regularization. 
We introduce these two algorithms  respectively in Sec.~\ref{section:MADQN} and
Sec.~\ref{section:SSMnet}. Then we provide the implementation details in Sec.~\ref{section:procedure}.

\subsection{Deep multi-agent  Q-learning}
\label{section:MADQN}

Deep Q-Network with single agent (DeepSaQ) has been proved to be effective in landmark detection with incomplete images \cite{ghesu2018towards}. 
By learning a navigation policy with local patches as the input, DeepSaQ can detect one target and tackle the problem of missing structures elegantly. 
When multiple targets are presented, multiple DeepSaQs need to be trained and deployed, 
which can result in deterring training time and memory demand due to its inefficiency.

We therefore propose the DeepMaQ, \textit{i.e.,} deep multi-agent  Q-learning network, which can efficiently detect dozens of landmarks at same time in presence of incomplete images. 
For each target landmark, an AI learner, \textit{i.e.,} an \textit{agent}, is defined. 
The navigation policy can be optimized by defining a dynamics function of Markov
decision process. 
For each frame, multi-series of local patches are constructed, as inputs, to guide the agents finding their paths to the corresponding target positions.
All agents share the same parameters and navigate simultaneously. 
Additionally, each agent can be independent of others. 
This reduces the unnecessary dependency of the \textit{effective agents}, when searching for their targets, on the \textit{ineffective agents} whose corresponding targets are not within the image due to the incompleteness of the image.
 

In multi-target RL, we assign one agent for the detection of one landmark, thus the set of indices for agents is denoted as $\mathcal{J}$, the same as that of target landmarks. 
The learning capability and efficiency  of a RL algorithm depend on the setting of state, action and reward of agents.
In the following, we elaborate on their settings in details.

\textbf{State}: In our model, the current observations (states) of agents are represented by a series of cropped patches from an image. 
Current state, denoted as $S$, is then defined as the set of current states of all agents, 
$\{s_{j} \vert _{j\in{\mathcal{J}}}\}$, 
of which each $s_{j}$, associated with agent $j$, is
a frame history buffer.
We define this buffer using the concatenation of patches centered on the positions of agent $j$ given the $n$ time frames, \textit{i.e.,} $s_j=\{\Gamma_{j}^{[t]}, \Gamma_{j}^{[t-1]}, ..., \Gamma_{j}^{[t-n+1]}\}$, where
$t$ and $n$ represent current time and time lag term, respectively.
The latest $n$ cropped patches are used together to stabilize the search trajectories. States $S$ are input to a deep neural network-based RL model (such as DeepMaQ), and the output decides which action should be taken for every agent.

\textbf{Action}: An agent takes a series of actions to navigate to target position. 
In our model, current action is the concatenation of actions  of all agents at current state, and is denoted as  $A=\{a_{j}\vert _{ j\in{\mathcal{J}}}\}$.
For simplicity, here we consider four types of actions for 2D images, including move left, move right, move up and move down with a preset step length, thus the action of agent $j$ from the action space is denoted as $a_{j} \in \{M_{\mathrm{left}},M_{\mathrm{right}},M_{\mathrm{up}},M_{\mathrm{down}}\}$. 
Similarly for 3D images, we further allow the agent to move forward and backward, denoted respectively by $\{M_{\mathrm{forward}}, M_{\mathrm{back}}\}$. 
In the implementation, we adopt the multi-scale strategy coupling step lengths with scales, namely the agents move in larger step lengths in larger scale settings. The moving length will decrease when the agent terminates at certain scale.  
An agent receives feedback from the environment with an action, and the feedback is implemented via the reward mechanism. 

\begin{figure}[t]
\centering
\includegraphics[width=0.95\textwidth]{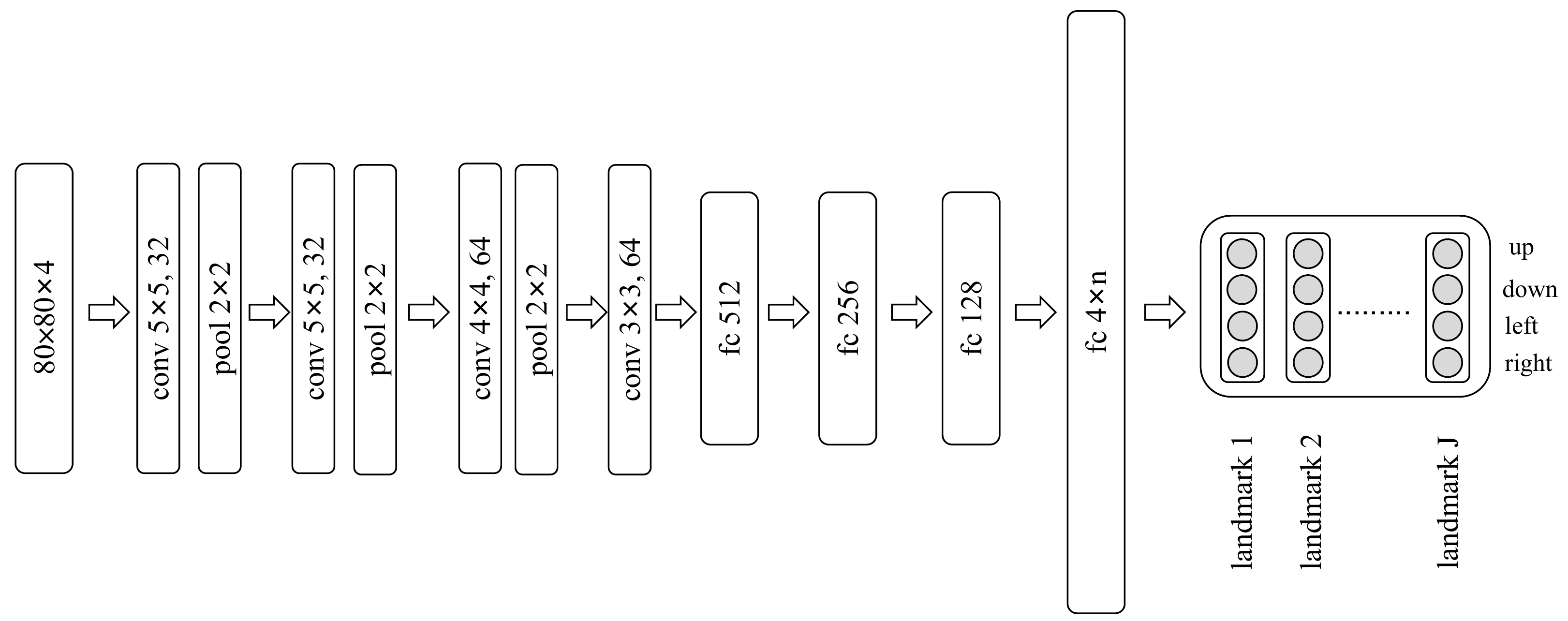}
\caption{\label{fig:fig3}Illustration of DeepMaQ network architecture.}
\end{figure}

\textbf{Reward}: At current state, an agent reaps a reward for taking an action, and different actions at different states of different agents receive different rewards.
We denote our reward function of multi-agent RL as a concatenation of individual rewards from each agent, \textit{i.e.,}   $R = \{r_{j}\vert _{j\in{\mathcal{J}}}\}$, 
where $r_{j} = D(x_j^d,x_j^{[t-1]})-D(x_j^d,x_j^{[t]})$, $x_j^d$ is the ground-true location for agent $j$, $x_j^{[t-1]}$ and $x_j^{[t]}$ are the locations of agent $j$ before and after taking an action, respectively. 
The measurement indicator $D$ used here is the Euclidean distance; $r_{j}$ is positive if distance between location and ground truth becomes smaller. 
The reward function is the foundation of the loss function in RL.

The action-selection policy in the proposed DeepMaQ can be optimized by learning Q-value function. Here, we approximate a Q-value function with a Deep Q-Network, \textit{i.e.,} the DeepMaQ-Net shown in Fig. \ref{fig:fig2}. Hence, the outputs of DeepMaQ-Net are the estimated Q values and measure the benefits of taking different actions given current state for the agents. 
Based on the Bellman Equation (\cite{ALANSARY2019156}), we propose the following loss function for our multi-agent RL, \textit{i.e.,} DeepMaQ, 
\begin{equation} \begin{array}{r@{}l} 
\mathcal{L}_{MaQ}\left(\omega\right)=  
  \displaystyle \sum_{i=1}^N\sum\limits_{j\in{\mathcal{J}_i}} &  \mathbb{E}_{s_j,a_j,r_j,s_j^{'}} \Big(r_j - Q_j(s_j,a_j;\omega) +  \\
    & \quad\quad\quad\quad  \gamma{\mathop{\max}_{a_j^{'}} Q_j }(s_j^{'},a_j^{'};\omega ) \Big)^{2}, 
\end{array}
\label{con:loss_madqn} \end{equation} 
where $s_j,\ a_j,$ and $r_j$ are current state, current action to take and immediate reward of taking this action for agent $j$, respectively; 
$s_j^{'}$ and $a_j^{'}$ are the new state and action to take after action $a_j$, respectively;
$Q_j()$ represents the Q-value function;   
$\mathop{\max}_{a_j^{'}} Q_j()$ denotes the maximal Q value for future actions to take, and $\gamma$ is the discount factor.

\begin{figure}[t]
\centering
\includegraphics[width=0.6\textwidth]{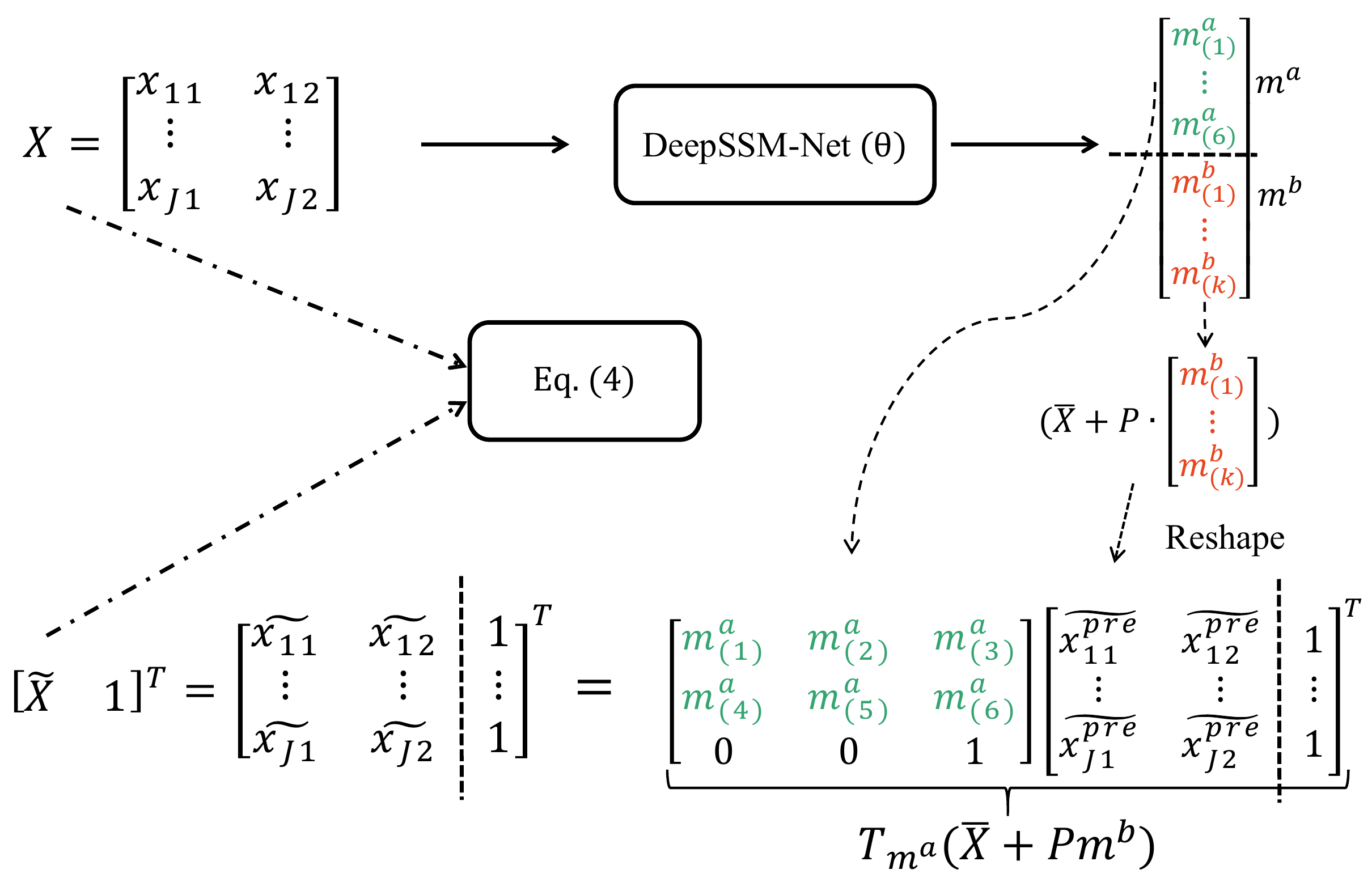}\\[-2ex] 
\caption{\label{fig:fig_ssm} Illustration of computations in the DeepSSM. $m_{(i)}^{a}(m_{(i)}^{b})$ represents the $i$th element of $m^{a}(m^{b})$.}
\end{figure}

\subsubsection{DeepMaQ network}
\label{section:deepmaq:network}

As Fig. \ref{fig:fig3} illustrates, DeepMaQ network is composed of four convolutional layers interleaved with three max-pool layers followed by four fully connected layers. The last layer was reshaped as a matrix according to the number of landmarks need to be detected. 
The $i$th element in the $j$th column of output matrix represents the utility values of executing the $i$th action for agent $j$, which shall decide $Q_j$ in Eq.~(\ref{con:loss_madqn}).
When agent $j$ needs to take an action, the corresponding state column is extracted as an input to the DeepMaQ-Net, as shown in Fig.~\ref{fig:fig2}.   
One can see that when the number of target landmarks increases, we only need to increase the number of parameters in the last two fully connected layers. Hence, this architecture is suitable for various tasks of multi-target landmark detection.

\subsection{Deep statistical shape model}
\label{section:SSMnet}
 
Multi-agent RL can predict a set of landmarks which however may form an unrealistic shape of the object.
This is attributed to the lack of global shape knowledge of the RL algorithm when no prior is provided. 
Hence, we introduce a statistical shape model (SSM) into the RL network, resulting in SGMaRL.  
To couple with the DeepMaQ, we further develop the deep neural network implemented SSM, referred to as the DeepSSM. 

Given an image, the detected landmark set, denoted as $X$, is aligned to a pre-constructed SSM, as follows,
\begin{equation}
\label{con:x+pb}
a^{*},\ b^{*}=\arg\min_{a,\ b}\Vert {X-T_{a}(\overline{X}+Pb)}\Vert ,
\end{equation}
where $\overline{X}$ represents the mean shape of the landmark set library for constructing the SSM; $T_a$ represents an affine transformation defining with vector $a$;
$P$ is the matrix formed from the largest $k$ modes of the SSM, and vector $b\in\mathbb{R}^{k}$ refers to weights to be optimized.
To find $a^{*}$ and $b^{*}$ which satisfy Eq.~(\ref{con:x+pb}), conventional methods generally adopt an iterative algorithm, which nevertheless can be time-consuming. 
We therefore propose to train a deep neural network of SSM (DeepSSM), whose model is denoted as $G$, to infer $a^{*}$ and $b^{*}$ in an efficient manner. The loss function of DeepSSM is given by,
\begin{equation}
\label{con:loss_ssm}
\mathcal{L}_{SSM}(\theta)= \sum_{i=1}^N\left({X_i-T_{{m^{a}_i}}(\overline{X}+Pm^{b}_i)}\right)^2,\ 
\end{equation}
where $[m^{a}_i,\ m^{b}_i]=G(X_i;\theta)$ are output of the DeepSSM. 

For the architecture, we adopt ResNet-18 \cite{7780459} as the backbone of DeepSSM. 
The output vector of $G$ is sliced into two sub-vectors, representing $m^{a}$ and $m^{b}$, respectively, as shown in Fig.~\ref{fig:fig_ssm}. For 2D images, $m^{a}$ has 6 elements and $m^{b}$ has $k$ elements;
 while for 3D images, $m^{a}$ has 12 elements and $m^{b}$ also has $k$ elements; $k$ is the number of modes used in the SSM.

\subsection{Implementation details}
\label{section:procedure}

\subsubsection{Training}
\label{subsubsection:Training process}
There are two deep neural networks in SGMaRL, \textit{i.e.,} DeepMaQ-Net and DeepSSM-Net, as Fig.~\ref{fig:fig2} illustrates.
The former is aimed to learn the navigation policy for every agent to find its target from a random starting position.
The latter is trained to map and regularize a random shape to the closest one in the shape space defined by the SSM.
The random shape here refers to the output of the DeepMaQ-Net and is defined by the set of detected landmarks, and the SSM is constructed using a pre-defined shape library. 
These two networks are trained alternately. Note that the outputs of half-way trained DeepMaQ-Net can provide useful samples for the training of DeepSSM-Net.
Alg.~\ref{algorithm_train} provides the pseudo code.

\textit{In the training of DeepMaQ-Net}, which is a RL-based neural network, we assign random initial positions for agents at the beginning of the episode for a training image. 
For these agents whose target landmarks are within this image, we consider them as being \textit{active} and allow them to take actions (navigate). 
By contrast, for those agents whose corresponding targets are in the missing areas, we consider them as being \textit{passive} and keep them fixed in the initial positions without any navigation. 
Similarly, when an agent reaches its own terminal state, we then assign it as being passive and keep it in the terminal position with no more action during the rest navigation steps for other active agents in this episode. 
This lasts until the end of the episode.

In addition, we adopt three techniques to ensure a stable and efficient training of DeepMaQ-Net, including the $\epsilon$-greedy \cite{ALANSARY2019156}, replay-and-freeze, and multi-scale strategies \cite{journal/mia/Zhuang2016}.
First, we propose to use $\epsilon$-greedy strategy, where the agent selects an action uniformly at random with probability $\epsilon$ in each step. 
This is because when the model is insufficiently trained, the agent can be misled by taking the action corresponding to the maximal Q value. Here, $\epsilon$ will be set as a smaller value when DeepMaQ becomes convergence. 
Second, experience replay memory and freezing the target network can be used to minimize the effects of instability and divergence in deep Q-learning networks. 
The former (replay) avoids the problem of successive data sampling, and the latter (freezing) helps reducing rapid changes in Q values. 
Finally, the multi-scale strategy with multiple resolution of images is adopted to improve the capture range and convergence speed.
Here, navigation step lengths are coupled with image resolution. Hence, navigation in course-scale images is with large physical step lengths and accelerates convergence towards the target plane; while navigation in fine-scale images steps in small length and fines tune the final estimation of plane parameter. The combination of multi-scale images contributes to better capture range and convergence speed.

\textit{Before training DeepSSM-Net}, we first employ a set of subjects represented by complete landmark sets to construct a SSM~\cite{cootes1995active},
with the results of mean shape $\overline{X}$ and mode matrix $P$ in Eq.~(\ref{con:x+pb}).
We denote this landmark set as $\mathbf{X}=\{X_i \vert_{i=1,\cdots, N}\}$.
Then, we generate random samples using resulting SSM for a pre-training of DeepSSM-Net, where the samples are generated using ${X^{*}}=T_{m_{*}^{a}}(\overline{X}+Pm_{*}^{b}))$. ${m_{*}^{a}}$ and $m_{*}^{b}$ are the gold standard parameters of $X^*$ for supervised learning. 
Finally, these generated samples can be added to $\mathbf{X}$ for the joint training of DeepMaQ-Net and DeepSSM-Net.


As Alg.~\ref{algorithm_train} shows, in the joint training the outputs of DeepMaQ-Net, which are inputs of DeepSSM-Net, are also samples for training the latter network. 
Hence, they are further included into the library $\textbf X$
for batch-based random sampling. 
Note that in the joint training, we use Eq.~(\ref{con:loss_ssm}) as the unsupervised loss function for DeepSSM-Net.

\begin{algorithm}[!t]
\caption{Training stage of SGMaRL}
\label{algorithm_train}
\KwData{Training set $\{(I_i, X_i,\mathcal{J}_i)\vert _{i=1,\cdots, N}\}$, number of training images $N$, landmark set library $\mathbf X$, 
maximum number of episodes $M$, budget $T$, greedy parameter $\epsilon$}
\KwResult{Optimal $\omega$, $\theta$}

Initialize DeepMaQ-Net with random weights $ \omega$\ , the lag weight $\omega^{-} = \omega$\ , replay memory $\mathbf D$ \;

Construct SSM, pre-train DeepSSM-Net and expand $\mathbf X$\;
\For{episode = 1, M }
{
Select a random image $I_i$\;
Initialise sequences $\{x_j\vert_{j \in \mathcal{J}_i}\}$ with random landmarks, state $\{s_{j}^{[0]}\vert_{j \in \mathcal{J}_i}\}$ according to $x_j$ \;
\For{t = 1, T}{
{If agent $\{j\vert_{j \in \mathcal{J}_i}\}$ is not terminated, performs $a_{j}^{[t]}$ according to DeepMaQ-Net with probability $1-\epsilon$, otherwise selects a random action\;

Get $x_{j}^{[t+1]}$, $r_{j}^{[t]}$, $s_{j}^{[t+1]}$ and store transition $(s_{j}^{[t]}, a_{j}^{[t]}, r_{j}^{[t]},s_{j}^{[t+1]})$ in $\mathbf D$\;

Sample random batch of transitions from $\mathbf D$, perform a gradient descent step on Eq.~( \ref{con:loss_madqn})
with respect to DeepMaQ-Net parameters $\omega$\;

Terminate agent $\{j\vert_{j \in \mathcal{J}_i}\}$ if it finds target, oscillates or reaches max steps, break when all agents are terminated\; 
}
}
Every $c_1$ steps, reset $\omega^{-} = \omega$ and update $\epsilon$\;

Every $c_2$ steps, test DeepMaQ-Net with $\{{I_i}^{c}\}$ from training set, get predictions $\{\hat{X_i}\}$ and store them in $\mathbf X$\;

Sample random batch from $ \mathbf X$, perform a gradient descent step on Eq.~(\ref{con:loss_ssm})
with respect to DeepSSM-Net parameters $\theta$\;
}
\end{algorithm}
\subsubsection{Test}
%
In the test stage, DeepMaQ-Net and DeepSSM-Net work in an interleaving and collaborative fashion, as the pseudo code in Alg.~\ref{algorithm_test} illustrates. 
After initialization of the agents, DeepMaQ-Net navigates all the agents with the actions predicted from the network, accordingly to the similar rules in the training stage. 
After the navigation ends, DeepSSM-Net then regularizes the set of predicted landmarks iteratively until the optimal solution of landmark positions are achieved or maximal number of subiteration steps ($T'$) is met. 
The resulting positions of landmarks are then fed into DeepMaQ-Net as the initial states of all the agents in DeepMaQ-Net for new round navigation. 
This interleaving application of DeepMaQ-Net and DeepSSM-Net last until the detection results converge or the iteration meets the maximal steps.

\begin{algorithm}[!t]
\caption{Test stage of SGMaRL}\label{algorithm_test}
\KwData{Test image set $\{I_{i}\vert _{i=1,\cdots, N^{'}}\}$, number of test images $N'$, 
index of all target landmarks $\mathcal{J}$, 
budget of DeepMaQ $T$, 
maximum number of subiterations with DeepSSM $T'$,
maximum number of iterations $M'$}
\KwResult{Landmarks prediction $\{X_i^{pred}\vert _{i=1,\cdots, N^{'}}\}$}

Select image $I_{i}$ from $\{I_{i}\vert _{i=1,\cdots, N^{'}}\}$\;
\For{iteration = 1, M'}{
Initialise sequence $\{x_j\vert_{j \in \mathcal{J}}\}$ with random landmarks if iteration = 1, otherwise initialise $\{x_j\vert_{j \in \mathcal{J}}\}$ with $X_i^{pred}$\;
\For{t = 1,T}{
If agent $\{j\vert_{j \in \mathcal{J}}\}$ is not terminated, performs $a_{j}^{[t]}$ according to DeepMaQ-Net ,moves from $x_{j}^{[t]}$ to $x_{j}^{[t+1]}$\;
 Terminate agent $\{j\vert_{j \in \mathcal{J}}\}$ if it oscillates or reaches max steps, update $x_j$ with $x_{j}^{[t+1]}$, break when all agents are terminated\;
}

{\For{t = 1,T'}{
Set $\Phi=\emptyset$\;
Add $\{j\vert_{j \in \mathcal{J}}\}$ to $\Phi$ if $x_j$ is judged to be corrected according to DeepSSM-Net, break if $\Phi=\emptyset$\;

Calculate alternative landmarks 
$\{\Tilde{x_j}\vert_{j \in \mathcal{J}}\}$ with DeepSSM-Net\;

Update $\{{x_j}\vert_{j \in \Phi}\}$ with $\{\Tilde{x_j}\vert_{j \in \Phi}\}$ if $\Tilde{x_j}$ is better than $x_j$ according to DeepSSM-Net\;

}
Update $X_i^{pred}$ with $\{{x_j}\vert_{j \in \mathcal{J}}\}$, break if $X_i^{pred}$ is convergence\;
}
}
\end{algorithm}
\section{Experiment} \label{exp}

\subsection{Incomplete image scenarios} \label{exp: imcomplete_scenario}
The proposed SGMaRL was evaluated in the following scenarios where incomplete images or labels could be encountered in either training or test stage, as illustrated in Fig.~\ref{fig:fig_intro}.
\begin{itemize}
    \item \textbf{(Baseline) Train: $I^{c}+L^{c}$, Target:
    $I^{c}\rightarrow L^{c}$}. The training set consists of \textbf{c}omplete images with \textbf{c}omplete labels,
    and the task is to predict \textbf{c}omplete labels on \textbf{c}omplete images. 
    
    \item \textbf{(Scenario I) Train: $I^{inc}+L^{inc}$, Target:
    $I^{c}\rightarrow L^{c}$}. The training set consists of \textbf{inc}omplete images with \textbf{inc}omplete labels, and the task is to predict \textbf{c}omplete labels  on \textbf{c}omplete images.
    
    \item \textbf{(Scenario II) Train: $I^{c}+L^{inc}$,
    Target:
    $I^{c}\rightarrow L^{c}$}. The training set consists of \textbf{c}omplete images with \textbf{inc}omplete labels, and the task is to predict \textbf{c}omplete labels  on \textbf{c}omplete images.
    
    \item \textbf{(Scenario III) Train: $I^{c}+L^{c}$, Target:
    $I^{inc}\rightarrow L^{inc}$}. The training set consists of \textbf{c}omplete images with \textbf{c}omplete labels, and the task is to predict \textbf{inc}omplete labels  on \textbf{inc}omplete images.
    
    \item \textbf{(Scenario IV) Train: $I^{c}+L^{c}$, Target:
    $I^{inc}\rightarrow L^{c}$}. The training set consists of \textbf{c}omplete images with \textbf{c}omplete labels, and the task is to predict \textbf{c}omplete labels  on \textbf{inc}omplete images.
\end{itemize}

For a given image, 
the missing landmarks are denoted as \textit{complementary label} ($L^{c-inc}$).
Noted that for complete images ($I^{c}$), it is possible that all landmarks are available and ${L^{c-inc}} = \emptyset$. But in some special cases, some landmarks may be absent, as shown in Fig. \ref{fig:exp2_data} (a).
For incomplete images ($I^{inc}$), ${L^{c-inc}} \neq \emptyset$. The landmarks near the edge of cropped images and those outsides belong to $L^{c-inc}$ (see green points in Fig. \ref{fig:exp1_data} (a)).

\subsection{Dataset}

\begin{figure}[thb]
\centering
\includegraphics[width=0.6\textwidth]{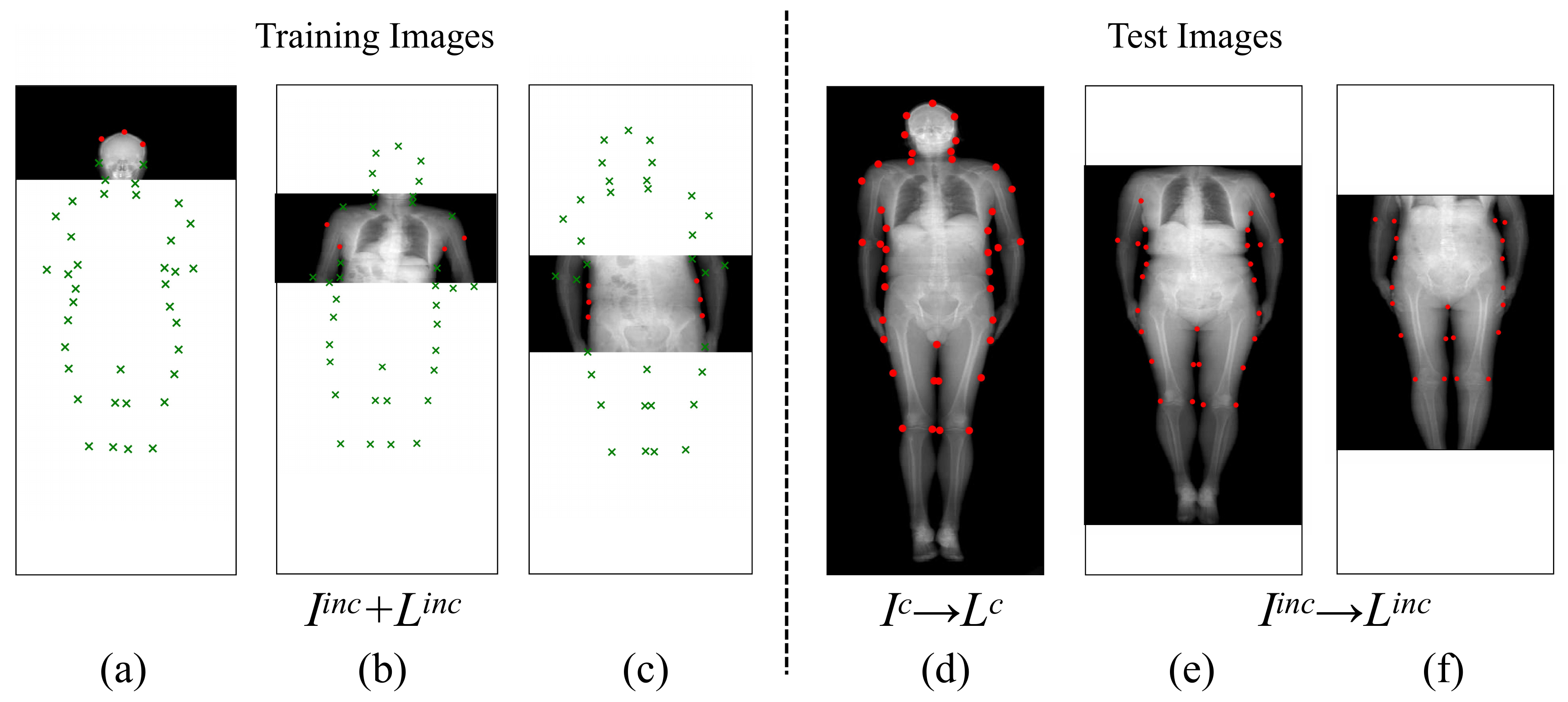}
\caption{\label{fig:exp1_data}Examples from the body dataset used in our experiments. (a), (b) and (c) represent training images while (d), (e) and (f) represent test images.
The dots (\textcolor{red}{$\bullet$})  represent landmarks known and the crosses
(\tiny\textcolor{green!50!black}{\XSolid} \footnotesize  )\ represent $L^{c-inc}$.}
\end{figure}

We used four datasets to validate the proposed method in the scenarios mentioned above.
The four datasets include (1) the whole-body dual energy X-ray absorptiometry (DXA) images, (2) the cardiac MRI images, (3) the whole-head CT images and (4) the half-head CT images.

The \textbf{whole-body DXA dataset} consists of 99 DXA original samples, provided by Zhongshan Hospital affiliated to Fudan University, Shanghai, China \cite{Gao2010TheSC}. The participants from Shanghai Changfeng Study were scanned over the whole body with lunar iDXA produced by GE company.
All these data were converted into images of JPG format with in-plane resolution of 3.0375 mm $\times$ 2.4001 mm.
Each sample was labelled with 40 contour keypoints by experts, as shown in Fig.~\ref{fig:exp1_data} (d). These landmarks can be used to describe the geometric shape of each individual.
This dataset was used for evaluation in Scenario I.

The \textbf{cardiac MRI dataset} consists of 45 LGE CMR sequences, collected from Shanghai Renji hospital \cite{10.1007/978-3-319-46723-8_67,8458220,zhuang2022cardiac}. 
Each sequence contains 10 to 18 slices with in-plane resolution of 1 mm $\times$ 1 mm, covering the main body of the ventricles. 
All 3D volumes were sliced and cropped to be 2D images with size of 256 $\times$ 256 in pixels.
The segmentation ground truth for left ventricle (LV), right ventricle (RV) and myocardium (Myo) were provided by experts. Only 2D slices containing all three areas were selected.
Segmentation labels on each image were preprocessed to generate 33 landmarks for model training, and at the inference stage the predicted landmarks would be transformed to deliver the segmentation output.
The details of transformation between segmentation and landmarks can be found in \ref{appendix_a}.
This dataset was used to evaluate the performance of the methods in Scenario II.

The \textbf{whole-head CT dataset} has 54 whole-head CT images from different participants, provided by Huashan Hospital affiliated to Fudan University, Shanghai, China \cite{QIAN2022934}. 
 Each volume was manually labeled by experts with 25 anatomical skull 
landmarks. These landmarks located on the surface of upper-half head, as shown in Fig. \ref{fig:exp3_data} (a).
This dataset was used to evaluate methods in Scenario III.

The \textbf{half-head CT dataset} consists of 41 upper-half head CT volumes from the Northern Han Chinese cohort, which was provided by Shanghai Institute of Nutrition and Health, Chinese Academy of Sciences.
The number of landmarks in these volumes varies from 14 to 19 according to the position of intersecting surface.
These volumes have much larger sample interval and lower resolution than samples from the whole-head dataset, as shown in Fig. \ref{fig:exp3_data} (a) and (b).
All of these volumes were reconstructed into an isotropic spacing of 1 mm along all three axes. This dataset was used for evaluation in Scenario IV.

\subsection{Implementation details}
In experiments, both DeepMaQ and DeepSSM were implemented on an NVIDIA GTX 1080Ti GPU.
For DeepMaQ, we adopted oscillation property, which means if an agent passes one location for several times, it should be terminated. 
The reward was set to be -1 to punish the agent when it moves across the border.
The experience replay memory size, the discount factor $\gamma$ and the time lag term were set as $10^5$, $0.9$ and $4$, respectively. 
For 2D and 3D networks of DeepMaQ, the sizes of input images were 80 $\times$ 80 and 30 $\times$ 30 $\times$ 30, respectively. The batch size was 48 for the former and 32 for the latter. Max steps for every agent was 200 for the former and 500 for the latter.
DeepMaQ took 24–36 hours and 48-72 hours for training 2D and 3D models, respectively.
For DeepSSM, we used complete labels for the construction of SSM and pre-training.
The maximum number of iterations $M'$ (see in \textbf{Alg.~\ref{algorithm_test}}) for test process was 5.
The pre-training time of DeepSSM was around an hour for both 2D and 3D models, respectively.

\subsection{Studies of \textbf{scenario I} using whole-body DXA}

$\textbf{Data preparation:}$
The body dataset was randomly divided into 40 and 59 images for training and test, respectively.
Each image in the training set was cropped horizontally with a predefined missing proportion ($mp$) $x\%$, which indicates that  $x\%$ image area is randomly removed, and the height of cropped image is $1-x\%$ of the original one.
To guarantee every agent learns the information of its corresponding landmark, each landmark index should be visited for at least once.
Examples can be found in Fig. \ref{fig:exp1_data} (a), (b) and (c).
To evaluate our model in Scenario I, the experiments were conducted in two settings:
(1) \textbf{Train:} {$I^{inc}+L^{inc}$}, \textbf{Target:} {$I^{c}\rightarrow L^{c}$}.
All images in the test set were complete, as shown in Fig. \ref{fig:exp1_data} (d).
(2) \textbf{Train:} {$I^{inc}+L^{inc}$}, \textbf{Target:} {$I^{inc}\rightarrow L^{inc}$}.
For training data, $mp$ was set to 0\%, 20\%, 40\%, 60\% and 80\%, respectively, while for test, it was set to 20\% and 40\% (see Fig. \ref{fig:exp1_data} (e) and (f)). 
In the following, we notated the missing proportion of training images and test images as \textit{$mp^{train}$} and \textit{$mp^{test}$}, respectively.
For comparison, we also implemented DeepMaQ, and a ResNet-based landmark detection method (denoted as ResNet) \cite{8687254}.
Euclidean distance was used as a distance metric and average distance error (ADE) was used to measure the evaluation accuracy of landmark detection.
 
$\mathbf{Results:}$
From Table \ref{tb:tb1}, one can see that SGMaRL performed well with training or test data being either complete or incomplete.
When testing on complete images, SGMaRL achieved 2.29 cm of ADE when $mp^{train}$ = 80\%, which was even better than the result of ResNet trained with complete images.
When testing on incomplete images, ResNet almost failed in all settings.
In contrast, when training with $mp^{train}$ being $80\%$, SGMaRL could still achieved 2.95 cm and 5.06 cm with $mp^{test} = 20\%$ and $mp^{test} = 40\%$, respectively.

\begin{figure}[!t]\center
\centering
\includegraphics[width=0.95\linewidth]{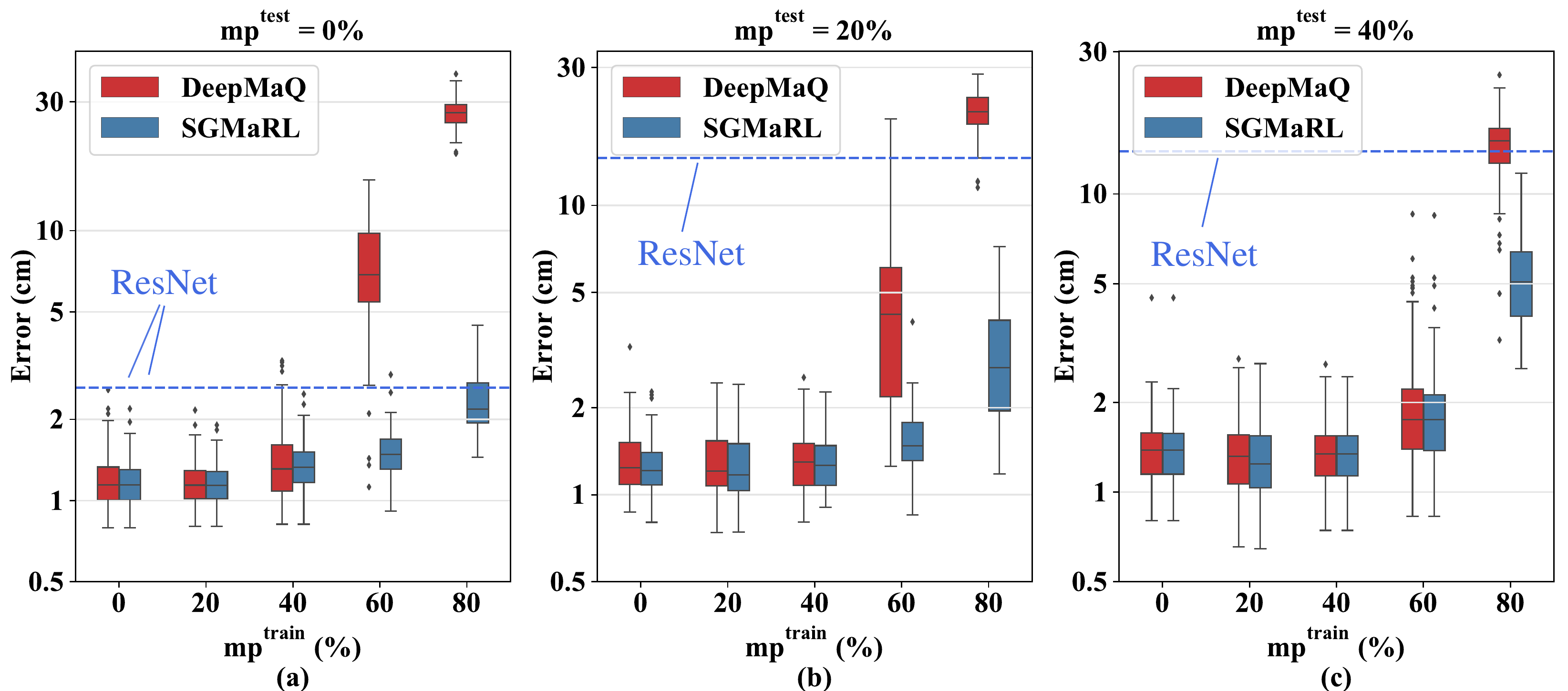}
\caption{\label{fig:exp_fig_1} Comparison between DeepMaQ and SGMaRL with different $mp$.
Predictions of ResNet with $mp^{train}$ = 0\% were used as baselines.}
\end{figure}

For better illustration, we further visualized these results in
Fig. \ref{fig:exp_fig_1}.
It shows that DeepMaQ and SGmaRL can both achieve acceptable results when $mp^{train}$ is small (e.g. $mp^{train}\leq 40\%$). 
Moreover, when $mp^{train}=80\%$, DeepMaQ failed while SGmaRL still worked.
As shown in Fig. \ref{fig:exp_fig_1} (a), when testing on complete images, there was no significant difference between predictions of DeepMaQ with $mp^{train} = 0\%$ and $mp^{train} = 20\%$ (p=0.733).
When $mp^{train}$ was less than 40$\%$, DeepSSM could only deliver marginal assistance, and ADE of both DeepMaQ and SGmaRL varied in a small range from 1.08 cm to 1.38 cm. 
However, when $mp^{train}$ rose from 60$\%$ to 80$\%$, ADE of DeepMaQ increased largely from 7.06 cm to 27.1 cm.
On the contrast, ADE of SGmaRL only increased slightly from 1.33 cm to 2.29 cm, thanks to the prior shape knowledge introduced by DeepSSM.
Testing on incomplete images showed a similar tendency (see Fig. \ref{fig:exp_fig_1} (b) and (c)).

\begin{table} 
\center
        \caption{Comparative results between DeepMaQ and SGMaRL with different $mp$. ResNet with $mp^{train}$ = 0\% was also included. ADE are in $cm$.}
\label{tb:tb1}
{\tiny
\begin{tabular}{c| lll|lll|lll}
\hline
\multirow{3}{*}{\makecell[c]{$mp^{train}$}}
&\multicolumn{9}{c}{$mp^{test}$}\\
\cline{2-10}
&\multicolumn{3}{c|}{0\% ($I^{c}$)}&\multicolumn{3}{c|}{20\%}&\multicolumn{3}{c}{40\%}\\

& DeepMaQ  & SGMaRL &  ResNet  & DeepMaQ  & SGMaRL &  ResNet&  DeepMaQ  & SGMaRL &  ResNet \\
\hline
0\% ($I^{c}$)   & 1.12$\pm$0.336   & 1.10$\pm$0.281   & 2.62$\pm$1.24 & 1.24$\pm$0.403 & 1.17$\pm$0.311 &14.6 $\pm$9.40 & 1.36$\pm$0.548& 1.35$\pm$0.532& 13.9$\pm$8.03 \\
20\%       & 1.10$\pm$0.274   & 1.08$\pm$0.244    & N/A & 1.20$\pm$0.350 & 1.18$\pm$0.324 & N/A & 1.27$\pm$0.415 &1.26$\pm$0.407 & N/A \\
40\%        & 1.38$\pm$0.593   & 1.30$\pm$0.339   & N/A & 1.25$\pm$0.365 & 1.22$\pm$0.316 & N/A & 1.28$\pm$ 0.353 & 1.27$\pm$0.336 & N/A\\
60\%         & 7.06$\pm$3.05  & 1.33$\pm$0.328   & N/A & 5.00$\pm$4.15 & 1.48$\pm$0.469  & N/A& 2.17$\pm$1.52 & 1.90$\pm$1.23  & N/A\\
80\%         & 27.1$\pm$3.79  & 2.29$\pm$0.703   & N/A & 21.0$\pm$3.62 & 2.95$\pm$1.72  & N/A& 14.3$\pm$4.08 & 5.06 $\pm$2.05& N/A \\
\hline
\end{tabular} }\\
\end{table}

\begin{figure}[!t]\center
\centering
\includegraphics[width=0.95\linewidth]{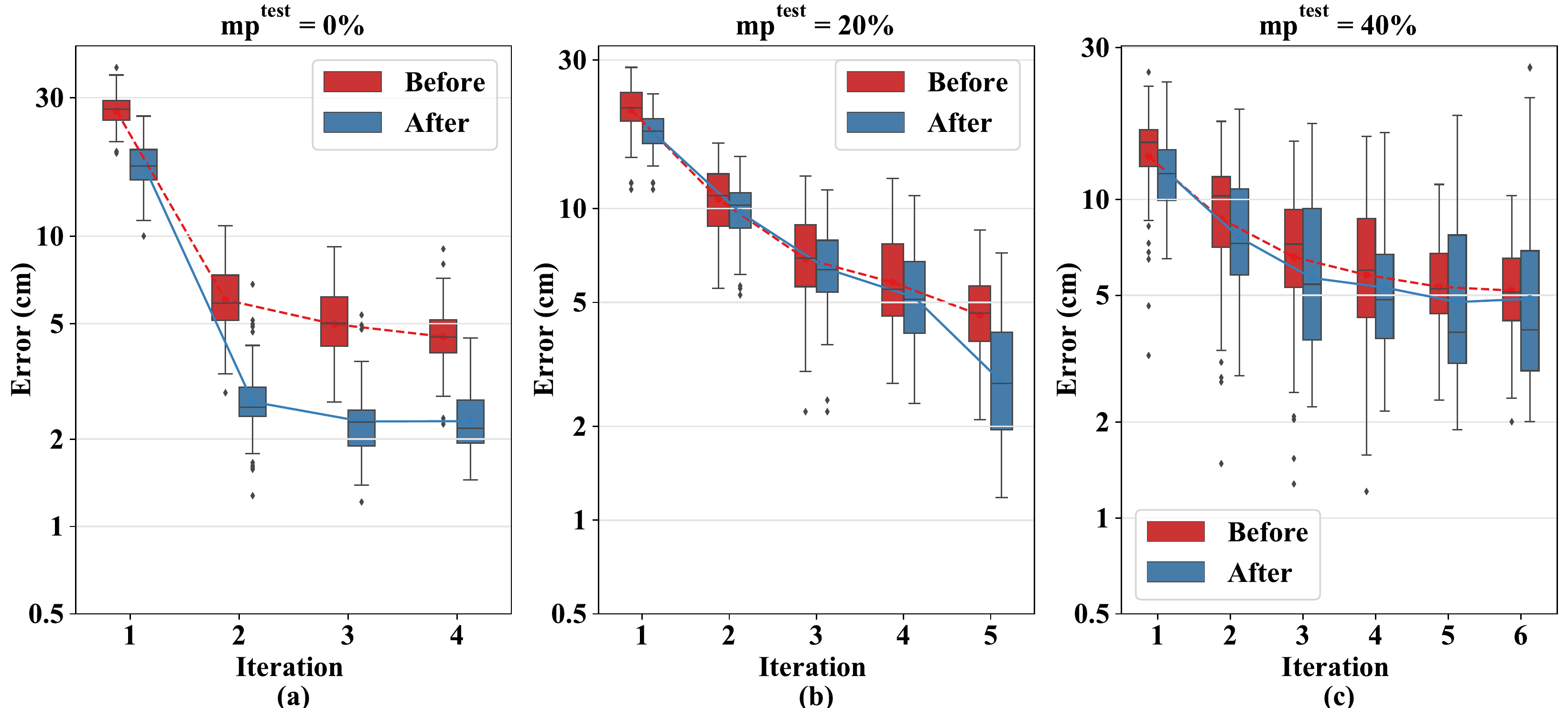}
\caption{\label{fig:exp_fig_2} Comparison before and after DeepSSM during iterations of SGMaRL with $mp^{train}$ = 80\%.
} 
\end{figure}

\begin{figure*}[!t]\center
\centering
\includegraphics[width=0.95\textwidth]{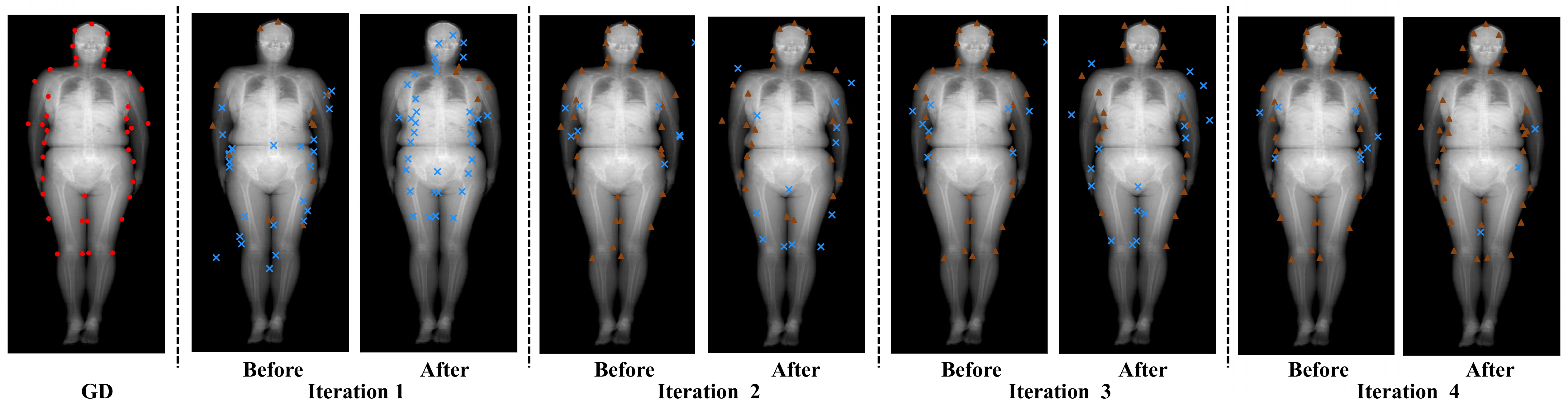}
\caption{\label{fig:exp_fig_3}The visualization of predictions before and after DeepSSM during iterations of SGMaRL with $mp^{train}$ = 80\%, $mp^{test}$ = 0\%.
The dots (\textcolor{red}{$\bullet$})  represent GD landmarks.
The triangles (\textcolor{brown!80!red}{$\blacktriangle$}) represent predictions of landmarks which are close to GD while the crosses (\tiny\textcolor{cyan}  {\XSolid}\footnotesize  ) represent abnormal predictions.
}
\end{figure*}

To show how DeepSSM helps during the test of SGMaRL,
we compared the predictions of SGMaRL before and after DeepSSM module in each iteration in
Fig. \ref{fig:exp_fig_2}.
As shown in Fig. \ref{fig:exp_fig_2} (b), before using DeepSSM, SGmaRL kept descending as the number of iterations increased. Besides, ADE of SGmaRL became smaller after adding DeepSSM in each iteration.
We conclude that DeepSSM can improve predictions both in each iteration and between iterations of SGmaRL in general.
Fig. \ref{fig:exp_fig_3} further provides visualization results. 
The predictions far from gold standards (GD) are denoted as \textit{abnormal predictions}. 
Abnormal predictions occurred at the beginning and affected the detection accuracy.
Moreover, as shown in Table \ref{tb:tb1} and Fig. \ref{fig:exp_fig_2}, most abnormal predictions were improved, with the number of iterations increasing. This is probably because DeepSSM provided sufficient shape prior information for landmark detection.
Besides, as shown in Table \ref{tb:tb1}, when $mp^{train}$ was $80\%$,
ADE of SGmaRL was 5.06 cm with $mp^{test} = 40\%$, which was much worse than 2.29 cm and 2.95 cm  with $mp^{test} = 0\%$ and $mp^{test} = 20\%$, respectively.
Additionally, when $mp^{test}$ = 40$\%$, Fig. \ref{fig:exp_fig_2} (c) showed larger variance and slower descent of performance during iterations, comparing to $mp^{test}$ = 0$\%$ and $mp^{test}$ = 20$\%$. 
It is probably because larger $mp^{test}$ with more $L^{c-inc}$ on test images could suppress the  advantages of DeepSSM.

\subsection{Studies of \textbf{scenario II} using  cardiac MRI}

\begin{figure}[!t]
\centering
\includegraphics[width=0.6\textwidth]{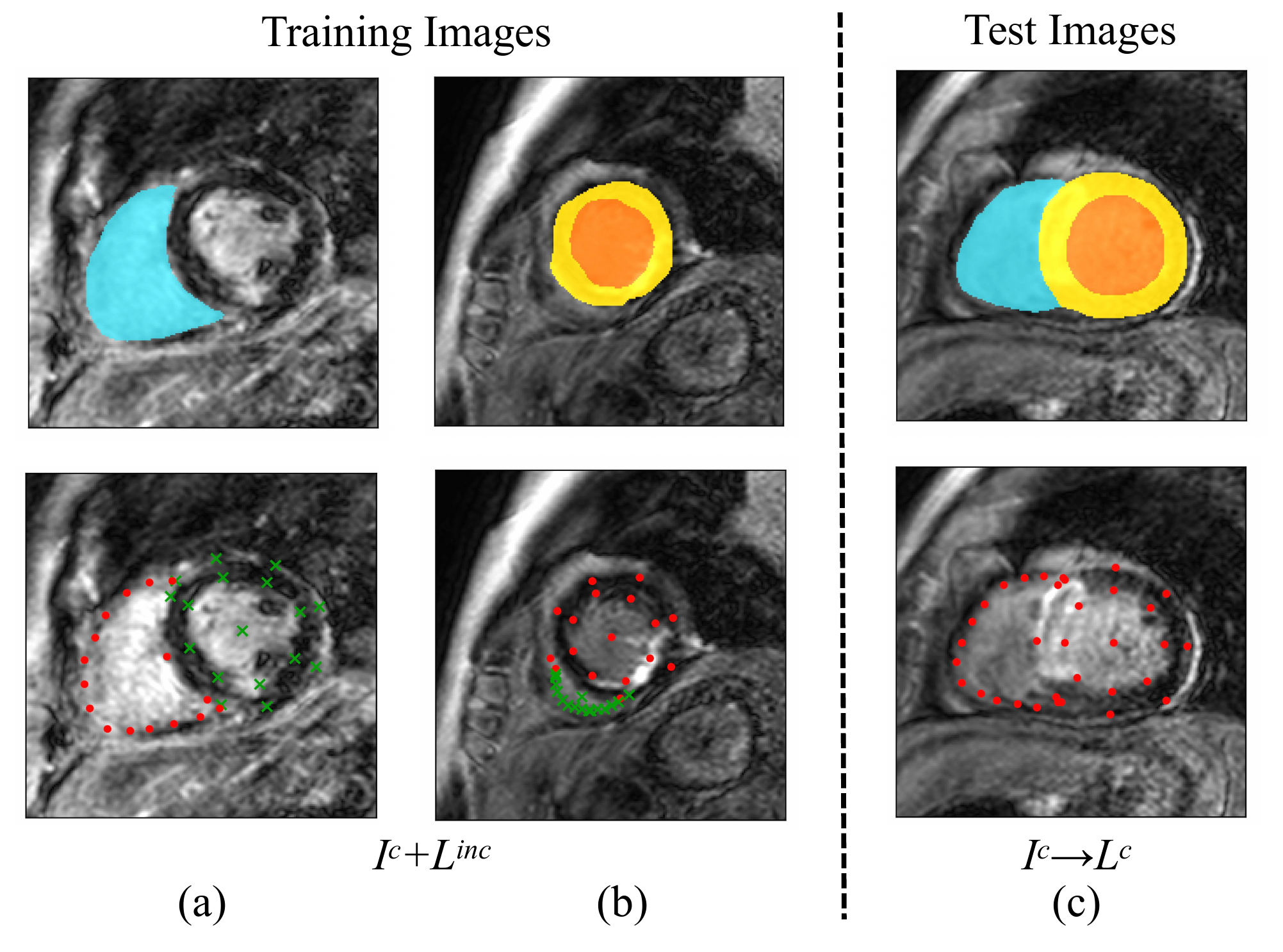}
\caption{\label{fig:exp2_data}Examples from the cardiac dataset used in our experiments. 
(a) and (b) represent training images while (c) represent test images. RV, LV and Myo are labeled in blue, orange and yellow, respectively.
The dots (\textcolor{red}{$\bullet$})  represent landmarks known and the crosses
(\tiny\textcolor{green!50!black}{\XSolid}\footnotesize  ) represent the 
$L^{c-inc}$.}
\end{figure}

$\mathbf{Data\ preparation:}$
The aim was to predict all labels (LV, RV and Myo) for complete images in test set. 
The final predicted labels were generated by DeepSSM based on the output of SGMaRL.
For comparison, the experiments were conducted in two settings:
(1) \textbf{Train:} {$I^{c}+L^{c}$, \textbf{Target:}
    $I^{c}\rightarrow L^{c}$}. The cardiac dataset was randomly divided into 20 and 25 samples for training and test, respectively. All training samples have labels of LV, Myo and RV. 
    ResNet and UNet were selected as landmark detection method and segmentation method for comparison, respectively.
    ResNet delivered the landmarks to segmentation output via post-processing, while UNet outputted segmentation prediction directly.
(2) \textbf{Train:} {$I^{c}+L^{inc}$, \textbf{Target:}
    $I^{c}\rightarrow L^{c}$}.  All training images were complete. However, half of the training images were labelled with only LV and Myo (see Fig. \ref{fig:exp2_data} (b)) while the others were labelled with only RV  (see Fig. \ref{fig:exp2_data} (a)).
    Because training with incomplete labels was more challenging, a little more samples were used in the training set.
    The cardiac dataset was randomly divided into 30 and 15 samples for training and test, respectively.
Dice score, average symmetric surface distance (ASD), Hausdorff distance (HD) and ADE were used as metrics for the evaluation.

\begin{table*} [!t]  
\center
\caption{Comparative results between ResNet, UNet, DeepMaQ and SGMaRL with different experiment settings.
}
\label{tb:tb3}
\resizebox{0.9\textwidth}{!}
{ 
\begin{tabular}{l|cll|cll|cll|c}
\hline
\multirow{2}{*}{Method} &\multicolumn{3}{c|}{LV}&\multicolumn{3}{c|}{Myo} &\multicolumn{3}{c|}{RV}&\multirow{2}{*}{ADE (mm)}\\
\cline{2-10}
&Dice&ASD (mm) &HD (mm) &Dice&ASD (mm) &HD (mm) &Dice&ASD (mm)&HD (mm) & \\
\hline
\multicolumn{3}{c}{\textbf{Train:} {$I^{c}+L^{c}$ \quad\textbf{Target:}
    $I^{c}\rightarrow L^{c}$}}\\
\hline
ResNet&	 
0.835$\pm$0.062&
3.70$\pm$1.37&	13.5$\pm$3.65&
	0.645$\pm$0.092&
	3.77$\pm$1.35&	13.6$\pm$4.23&
		0.774$\pm$0.074&	3.93$\pm$1.23&	16.1$\pm$3.02&7.67$\pm$
3.46\\
UNet&0.910$\pm$0.045&
2.49$\pm$1.35&	29.5$\pm$16.1&
	0.816$\pm$0.067&
	2.11$\pm$1.03&	13.4$\pm$10.0&	0.845$\pm$0.197	&3.69$\pm$2.75&	54.3$\pm$24.8& N/A\\

DeepMaQ& 0.898$\pm$0.045&	
2.59$\pm$0.851&	15.9$\pm$8.49&	0.754$\pm$0.056&
2.26$\pm$0.783&	14.9$\pm$10.5&0.833$\pm$0.085&	2.91$\pm$1.23&	19.0$\pm$7.51&
7.78$\pm$4.16
 \\
SGMaRL & 0.875$\pm$0.045&
3.09$\pm$0.732&	12.5$\pm$5.84&0.702$\pm$0.064&
2.92$\pm$0.888&	11.7$\pm$6.94&	0.778$\pm$0.082&	3.72$\pm$1.23&	16.2$\pm$5.67&
7.40$\pm$
4.03\\
\hline
\multicolumn{3}{c}{\textbf{Train:} {$I^{c}+L^{inc}$ \quad\textbf{Target:}
    $I^{c}\rightarrow L^{c}$}}\\
\hline
DeepMaQ&0.840$\pm$0.074& 3.82$\pm$1.47& 18.1$\pm$7.57& 0.650$\pm$0.100&5.29$\pm$1.51&16.2$\pm$6.77
& 0.780$\pm$0.073&3.78$\pm$1.20 & 17.3$\pm$7.60& 8.66$\pm$4.79\\
SGMaRL& 0.861$\pm$0.065 &3.04$\pm$1.30& 12.1$\pm$5.43& 0.693$\pm$0.091&3.12$\pm$1.32&12.0$\pm$5.38
 & 0.825$\pm$0.086& 3.89$\pm$1.35 & 14.1$\pm$4.53&8.19$\pm$4.65\\
\hline
\end{tabular}}\\
\end{table*}


\begin{figure*}[!t] 
\centering
\includegraphics[width=0.98\linewidth]{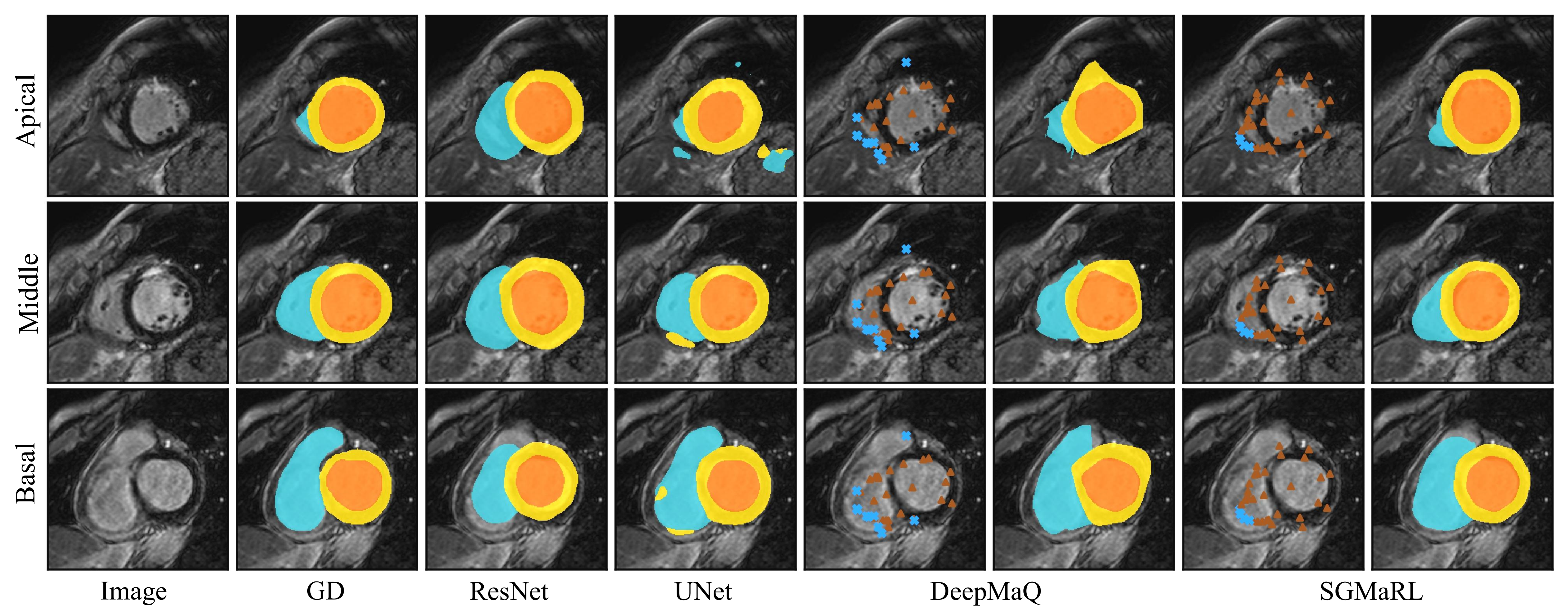}
\caption{\label{fig:exp3_fig_1}Comparison of prediction results of ResNet, UNet, DeepMaQ and SGMaRL with GD in three typical slices.
RV, LV and Myo are labeled in blue, orange and yellow, respectively.
The triangles (\textcolor{brown!80!red}{$\blacktriangle$}) represent predictions of landmarks which are close to GD while the crosses ({\tiny\textcolor{cyan}{\XSolid}}) represent abnormal predictions.}
\end{figure*}

$\mathbf{Results:}$
Table \ref{tb:tb3} shows that SGMaRL can successfully predict landmarks and segmentation when training with complete or incomplete labels.
For landmark detection, 
SGMaRL could reach 7.40 mm  of ADE with complete labels, while ResNet reached 7.67 mm.
With incomplete labels, SGMaRL could still achieve 8.19 mm of ADE.
For segmentation, SGMaRL also had advantage over ResNet for most metrics, but there was no significant difference between HD of RV (p = 0.807).
It also had great advantage over UNet in terms of HD.
SGMaRL could reach 16.2 mm in terms of HD for RV, while UNet only could reach 54.3 mm.
Note that UNet obtained better Dice scores and ASD compared to SGMaRL. 
This is because the segmentation obtained by SGMaRL was converted from predicted landmarks, which may introduce additional errors compared to the method that directly predicted pixel-wise segmentation.

To show the advantages of SGMaRL in preserving shape, we compared results of different methods and visualized them in Fig. \ref{fig:exp3_fig_1}.
It shows that SGmaRL could predict smooth and accurate shape.
In contrast, 
without structural information provided by DeepSSM, the prediction of DeepMaQ had irregular shape, due to the existence of abnormal landmarks prediction.
The prediction of UNet might have an irregular boundary and many noises, which caused large HD (see Table \ref{tb:tb3}).
ResNet could predict a reasonable shape, but the prediction might be sometimes inaccuracy, which caused low Dice scores.
Fig. \ref{fig:exp3_fig_1} also shows SGMaRL can deal with variety of shapes effectively. 
In different slices, the shape of RV expressed complex variation.
With the help of DeepSSM, SGMaRL can correct the abnormal landmarks and make prediction reasonable.

Table \ref{tb:tb3} further shows SGMaRL has additional advantages when training with incomplete labels. For example, the Dice score of RV rises from 0.780 to 0.825 after adding the DeepSSM.
Besides, HD averagely decreased by 3.13 mm and 4.47 mm when training with complete and incomplete labels, respectively.
It is possibly because training with incomplete labels could be quite challenging for DeepMaQ. 
With worse initial prediction, DeepSSM can improve HD more effectively.
Moreover, SGMaRL has better segmentation performance in terms of contours than filled regions.
For example, when training with complete labels, 
HD of all areas were improved with the assistance of DeepSSM. 
However, the Dice score of LV was 0.898 with DeepMaQ, while it dropped 0.023 with SGMaRL.
We concluded that DeepSSM can improve HD and ADE effectively, but its influence on Dice score and ASD still needs further research.

\subsection{Studies of \textbf{scenario III and IV} using head CT}

\begin{figure}[!t]
\centering
\includegraphics[width=0.6\textwidth]{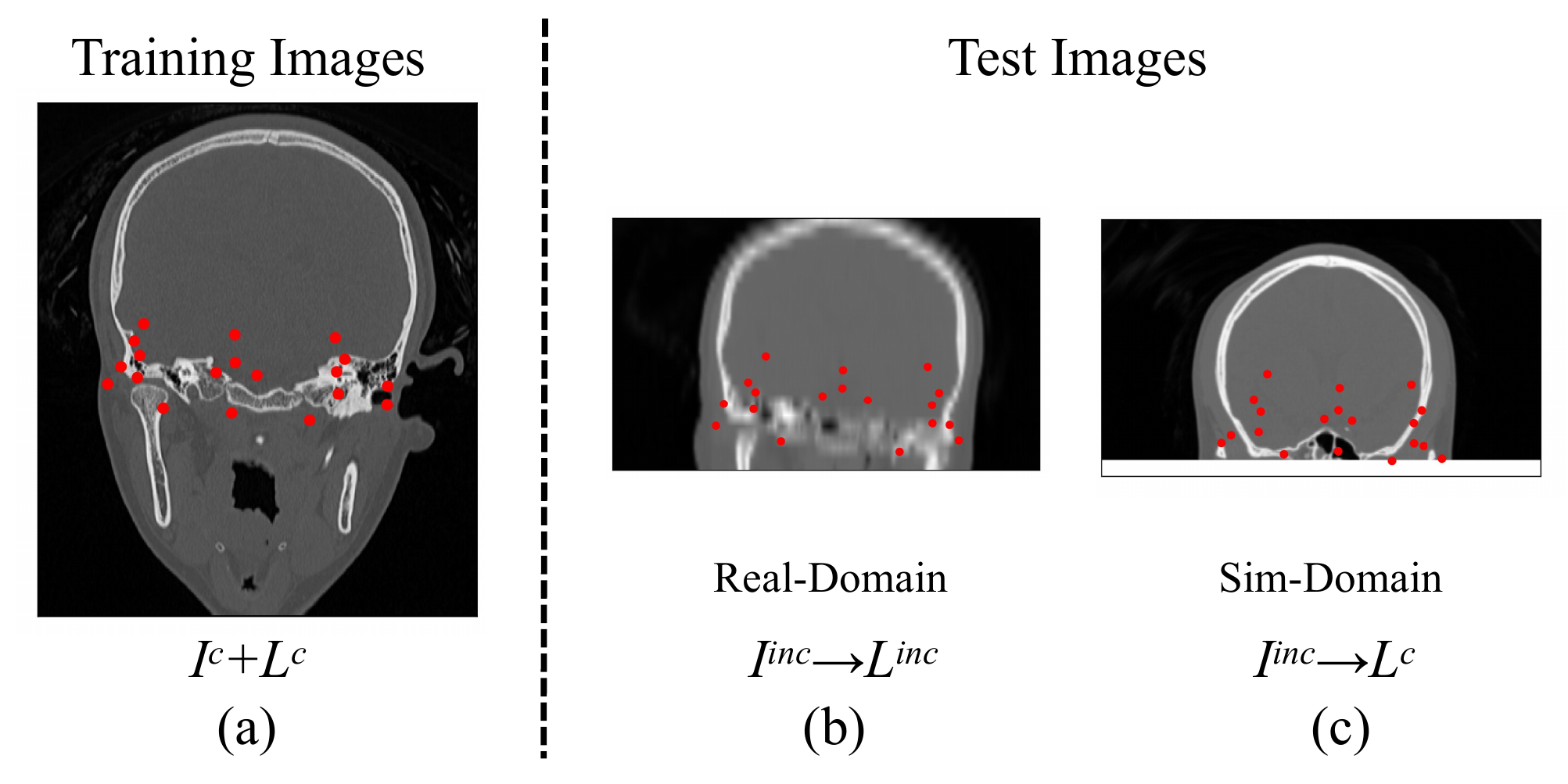}
\caption{\label{fig:exp3_data}Examples from the head datasets used in our experiments. 
(a) represents training images while (b) and (c) represent test images.
The dots (\textcolor{red}{$\bullet$})  represent landmarks known.
Note that all images in this figure are 2D slices of 3D images from front view, and landmarks on them are mapped from 3D space. 3D head images are used in our experiment.}
\end{figure}

$\mathbf{Data\ preparation:}$
We evaluated our methods in Scenario III and IV and conducted the experiments in two settings:
(1) \textbf{Train:} {$I^{c}+L^{c}$, \textbf{Target:}
    $I^{inc}\rightarrow L^{inc}$}.
We randomly selected 30 whole-head images from the whole-head dataset for training, while obtained 41 half-head images  from the half-head dataset for test. The test set was denoted as \textbf{Real-Domain}, as shown in Fig. \ref{fig:exp3_data} (b).
The number of landmarks labeled by experts on image from Real-Domain does not exceed 19.
(2) \textbf{Train:} {$I^{c}+L^{c}$, \textbf{Target:}
    $I^{inc}\rightarrow L^{c}$}.
The training set is the same as setting (1), while for the test set, other 24 whole-head images were half cut as simulated upper-half head images. This test set was denoted as \textbf{Sim-Domain},
as shown in Fig. \ref{fig:exp3_data} (c).
To ensure the high similarity between Sim-Domain and Real-Domain, we varied the landmarks in training data of Sim-Domain from 14 to 19 via controlling the position of intersecting surface when half cutting the whole images.
For setting (1), $L^{inc}$ (landmarks labelled by expert) were considered for evaluation.
For setting (2), we predicted all existed landmarks, which were divided into $L^{inc}$ (landmarks inside of cropped images) and $L^{c-inc}$ (landmarks near the edge of cropped images or those outsides).
Landmark detection methods based on multi-atlas registration (MAS) and single-atlas registration (SAS) from \cite{5444972} were used for comparison.
Both of them were implemented on an Intel Core i7-5700HQ CPU.
ADE was used for evaluation of landmarks detection accuracy.

\begin{table} [!t] \center
\caption{Comparative results MAS, SAS, DeepMaQ and SGMaRL with different types of landmarks in different domains. 
Noted inference time of MAS and SAS were observed when running on CPU.
ADE are in $mm$.
 } \label{tb:tb2}
\resizebox{0.8\linewidth}{!}{
\begin{tabular}{l| c|cc|c}
\hline
\multirow{2}{*}{Method}      & 
    Real-Domain  & \multicolumn{2}{c|}{ 
    Sim-Domain} &\multirow{2}{*}{Inference Time } \\
&$L^{inc}$
 & $L^{inc}$ & $L^{c-inc}$&  \\
\hline
MAS               
&6.10$\pm$1.59 
& 2.87$\pm$0.782  & 
6.80$\pm$0.867  &   
35 min 30 sec 
\\
SAS  &   
\multicolumn{1}{c|}{7.36$\pm$1.50}& 5.58 $\pm$1.97  & 21.8$\pm$10.1&  
1 min 11 sec\\   
DeepMaQ & \multicolumn{1}{c|}{8.99$\pm$2.84}                & 5.12$\pm$1.85     & 11.4$\pm$7.18       & 
13 sec\\ 
SGMaRL  & \multicolumn{1}{c|}{7.25$\pm$1.11}         & 5.01$\pm$1.83     & 6.84$\pm$4.82     &   
13 sec\\ 
    
\hline
\end{tabular} }\\
\end{table}

$\mathbf{Results:}$
As shown in Table \ref{tb:tb2}, our proposed method can successfully find different types of landmarks in both domains. The result of SGMaRL was better than SAS but worse than MAS. For example, ADE of $L^{inc}$ with SGMaRL was 7.25 mm in Real-Domain, which was smaller than 7.36 mm by SAS, but larger than 6.10 mm by MAS. However, SGMaRL consumed far less time than MAS. MAS took more than 35 min for one case with CPU while SGMaRL only took about 13 sec with GPU.

For most methods, the prediction of $L^{inc}$ in Real-Domain was harder than $L^{inc}$ in Sim-Domain, while $L^{c-inc}$ in Sim-Domain was the most difficult.
For example, DeepMaQ achieved ADE of 5.12 mm, 8.99 mm and 11.4 mm on $L^{inc}$ in Sim-Domain, $L^{inc}$ in Real-Domain and $L^{c-inc}$ in Sim-Domain, respectively.
Fig. \ref{fig:exp3_data} shows Real-domain test set has obvious appearance gap with the training images and poorer image quality, which increases the difficulty for detection task.
For $L^{c-inc}$, it could be only inferred with structural information since their local appearance information was almost missing.
With the help of DeepSSM, ADE of SGMaRL decreased by 0.11 mm, 1.74 mm, 4.56 mm in these landmarks, respectively.
We concluded that DeepMaQ has worse prediction when target image has poor appearance information.

\begin{figure}[!t]
\centering
\includegraphics[width=0.6\textwidth]{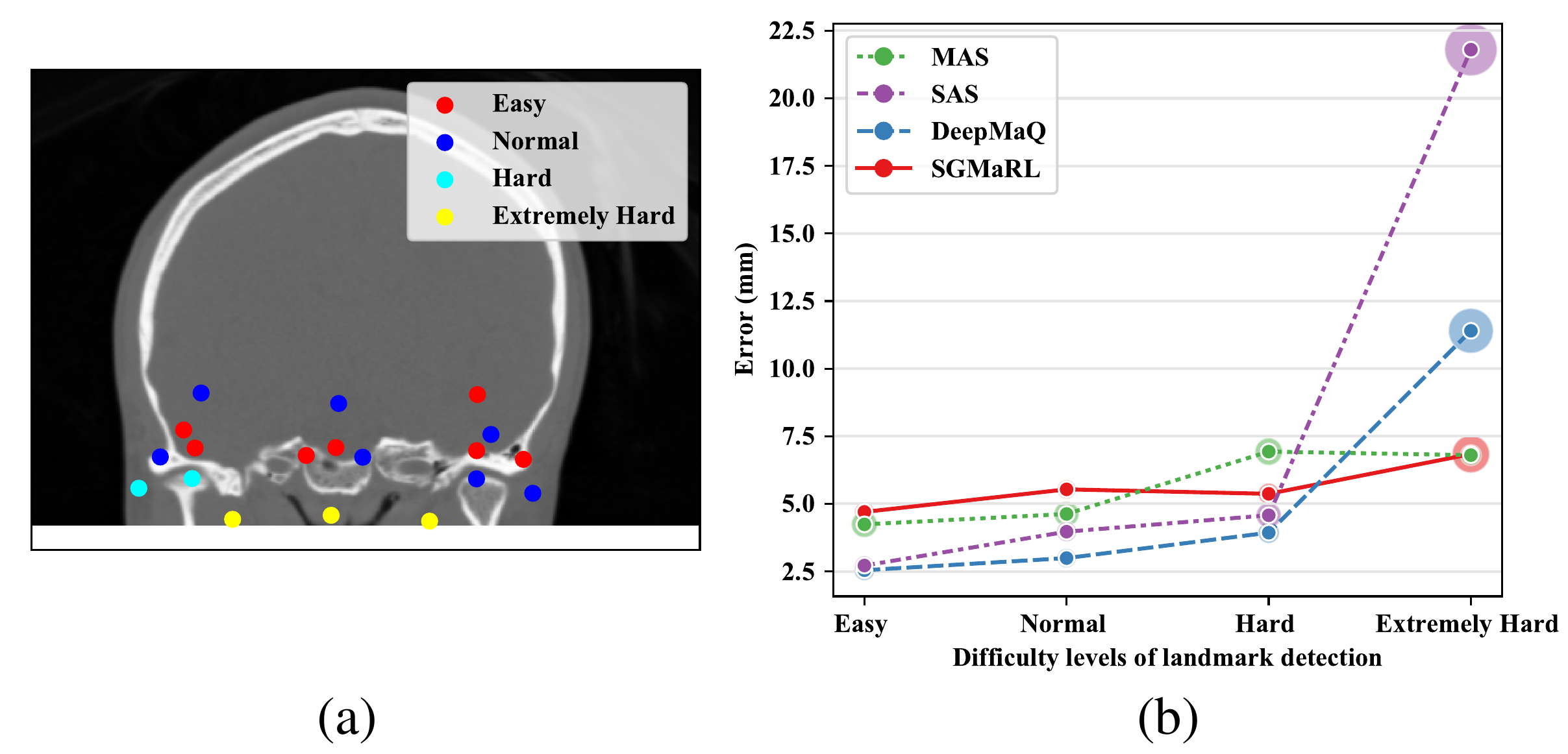}
\caption{\label{fig:exp2_fig_1} The right graph shows comparative result of DeepMaQ, SGMaRL, MAS, and SAS on landmarks with different difficulty levels in Sim-Domain. The size of bubble represents the standard deviation. The left graph shows how different difficulty levels of landmarks located on a head image. Extremely hard represents landmarks which belong to $L^{c-inc}$ on every image, while Hard represents the landmarks which belong to $L^{c-inc}$ on this image but belong to $L^{inc}$ on other images. The rest existent landmarks were sub-divided into Easy and Normal according to inter-observer errors.}
\end{figure}

To investigate the robustness of the proposed method, we further divided $L^{inc}$ and $L^{c-inc}$ in Sim-domain into four categories, i.e., Easy, Normal, Hard and Extremely Hard, according to their detection difficulties and compared the performance on them.
Fig. \ref{fig:exp2_fig_1} (a) illustrates the classification of the landmarks, and (b) shows the performance of the compared methods in the four categories.
One can see that from Easy to Hard, SGMaRL had a stable performance, while ADE of other three methods increased. When tested on Extremely Hard samples, DeepMaQ and SAS performed poor, while the prediction error of SGMaRL only increased slightly, when compared to the results on Hard samples. The results show that SGMaRL has great robustness on landmarks with different levels of detection difficulty.


\section{Conclusion and discussion}\label{conclusion}
In this paper, we proposed a shape-guided multi-agent RL algorithm for landmark detection with incomplete images. 
The algorithm contains two modules: DeepMaQ and DeepSSM.
The former can detect multiple landmarks as targets synchronously with incomplete images for training;
the latter can correct abnormal predictions and provide more accurate start positions for DeepMaQ. 
By combining them, the proposed method can exploit both local image information and shape information effectively, which is fast and robust for solving the general problem of multiple landmark detection with challenging incomplete images.
Experiments conducted on different datasets show that our methods can perform landmark detection in different incomplete image problems and complex scenarios.

We have investigated the performance of our methods upon different $mp$ in both training and test set.
The results show that DeepSSM can achieve greater improvement when $mp^{train}$ increases.
We also have carried out comparison study of the prediction errors of our methods with different types of landmarks.
The results show that (1) with prior shape information from DeepSSM, SGMaRL can even predict the landmarks outside the target image;
(2)
SGMaRL is robust on landmarks with different detection difficulties. Besides, we have extended our methods to segmentation task. The results show that DeepSSM can ensure the segmentation prediction of SGMaRL have a smooth shape based on landmarks as partial labels for training.

Though DeepSSM can provide great help, it can be explored for further improvement.
As mentioned before, the assistance of DeepSSM became weaker as $mp^{test}$ increasing. The details of DeepSSM 
are to be improved and refined to find abnormal landmarks more accurately, especially when $mp^{test}$ is large.
When performing segmentation, Dice score and ASD became worse after adding DeepSSM with complete labels for training.
We will investigate the influence of DeepSSM on different metrics and hope modified DeepSSM can improve all metrics in different scenarios.
Besides, learning the search strategy and localization in two stages may increase the possibility of abnormal predictions.
Thus, it will be interesting to investigate a more elegant way to combine DeepMaQ with prior shape information.
For this purpose, a new multi-agent RL sharing structural information with each agent will be studied in the future.

\end{spacing}

\section*{Acknowledgment}
This work was supported by the National Natural Science Foundation of China (61971142, 62111530195 and 62011540404) and the development fund for Shanghai talents (2020015).

\appendix
\section{Transformation between segmentation labels and landmarks for cardiac MRI dataset}\label{appendix_a}

\begin{figure}[h]
\centering
\includegraphics[width=0.95\textwidth]{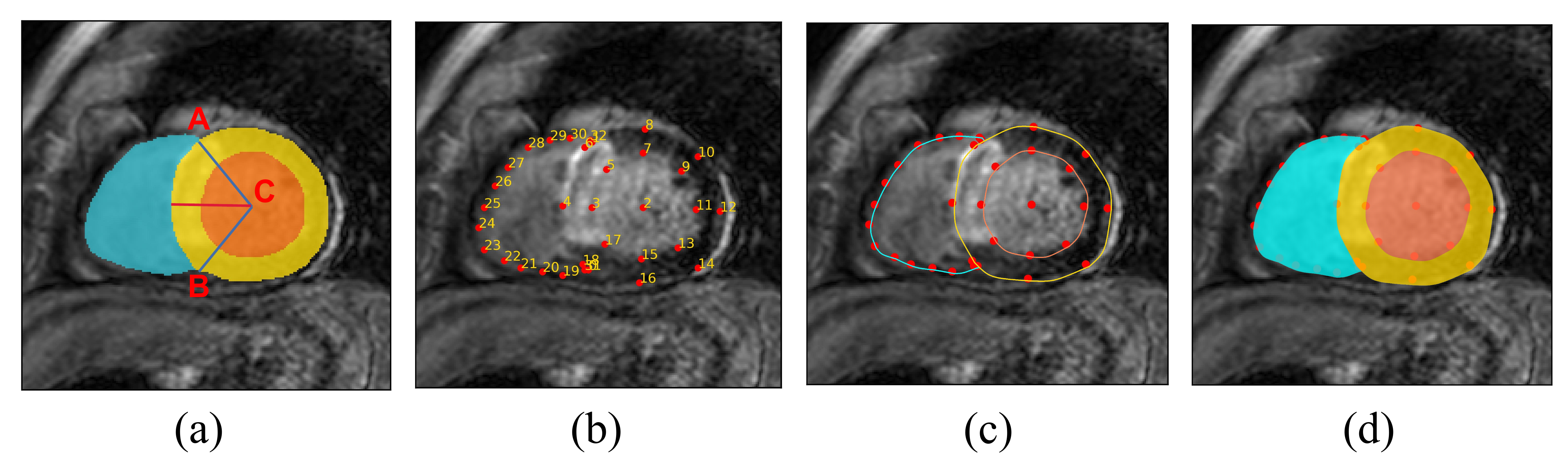}
\caption{\label{fig:fig_app} Illustration of the translation between segmentation label and landmarks. Landmarks in (b) are generated from segmentation in (a). 
The contour lines of (c) are generated from (b) and then transformed into final segmentation in (d).}
\end{figure}
Segmentation of every cardiac image was transformed into 33 landmarks, according to the following process.
First, the two intersection points of RV, Myo and background regions were denoted as point $A$ and point $B$, respectively. The center of LV was denoted as point $C$, as shown in Fig. \ref{fig:fig_app} (a). The three points were also the landmarks with index $0 \sim 2$ in Fig. \ref{fig:fig_app} (b).
From point $C$, we drew a ray in the bisector of $\angle ACB$. The intersection points of this ray with endosurface and episurface of LV were marked.
We rotated the ray around point $C$ at $45^{\circ}$ for 7 times, and obtained landmarks with index $3 \sim 18$.
In the end, 14 points were evenly distributed on the contour of RV, which were the landmarks with  index $19 \sim 32$.

The predicted landmarks were transformed into segmentation, according to the following process. 
First, we connected the related landmarks with Bézier curves, and drew the contour of RV, LV and Myo, as shown in Fig. \ref{fig:fig_app} (c). Then we filled the corresponding area as shown in Fig. \ref{fig:fig_app} (d).

To make sure the transformation can be performed correctly, the selected 2D slices should satisfy certain conditions: 1. There should be no space between Myo and RV, otherwise the landmark on the border of Myo and RV (landmark with index 4 in Fig. \ref{fig:fig_app} (b)) can not be generated  correctly.
2. Area of RV cannot be too small, otherwise landmarks on the contour of RV (landmarks with index $19 \sim 32$ in Fig. \ref{fig:fig_app} (b)) may overlap.
3. RV must be an integral area, otherwise
the landmarks on the contour of RV can not be generated to segmentation correctly.

\end{document}